\def\etal{\textit{et~al.}}
\def\etc{\textit{etc.}}
\def\ie{\textit{i.e.}}
\def\eg{\textit{e.g.}}
\def\wrt{\textit{w.r.t. }}
\def\vs{\textit{v.s. }}

\documentclass[10pt,journal,compsoc]{IEEEtran}

\ifCLASSOPTIONcompsoc
  \usepackage[nocompress]{cite}
\else
  \usepackage{cite}
\fi
\ifCLASSINFOpdf
   \usepackage[pdftex]{graphicx}
  \graphicspath{{../pdf/}{../jpeg/}}
  \DeclareGraphicsExtensions{.pdf,.jpeg,.png}
\else
  \usepackage[dvips]{graphicx}
  \graphicspath{{../eps/}}
  \DeclareGraphicsExtensions{.eps}
\fi
\newcommand\MYhyperrefoptions{bookmarks=true,bookmarksnumbered=true,
pdfpagemode={UseOutlines},plainpages=false,pdfpagelabels=true,
colorlinks=true,linkcolor={black},citecolor={black},urlcolor={black},
pdftitle={Bare Demo of IEEEtran.cls for Computer Society Journals},
pdfsubject={Typesetting},
pdfauthor={Michael D. Shell},
pdfkeywords={Computer Society, IEEEtran, journal, LaTeX, paper,
             template}}
\ifCLASSINFOpdf
\usepackage[\MYhyperrefoptions,pdftex]{hyperref}
\else
\usepackage[\MYhyperrefoptions,breaklinks=true,dvips]{hyperref}
\usepackage{breakurl}
\fi

\usepackage{booktabs}
\usepackage{bbding}
\usepackage{multirow}
\usepackage{multicol}
\usepackage[para,online,flushleft]{threeparttable}

\usepackage{amsmath}
\usepackage{amssymb}
\usepackage{array}
\usepackage{bbm}
\usepackage{calc}
\usepackage{caption}
\usepackage{color}
\usepackage{enumitem}
\usepackage{epsfig}
\usepackage{epstopdf}
\usepackage{graphicx}
\usepackage{hyperref}
\usepackage{multirow}
\usepackage{subfigure}
\usepackage{booktabs}
\usepackage{url}
\usepackage{xspace}
\usepackage[dvipsnames]{xcolor}
\usepackage{bm}
\usepackage{dsfont}

\newcommand{\figref}[1]{Fig\onedot~\ref{#1}}
\newcommand{\equref}[1]{Eq\onedot~\eqref{#1}}
\newcommand{\secref}[1]{Sec\onedot~\ref{#1}}
\newcommand{\tabref}[1]{Tab\onedot~\ref{#1}}

\def\onedot{\ifx\@let@token.\else.\null\fi\xspace}
\def\eg{\emph{e.g}\onedot} 
\def\ie{\emph{i.e}\onedot} 
 
\def\etc{\emph{etc}\onedot} \def\vs{\emph{vs}\onedot}
\def\wrt{w.r.t\onedot} 
\def\etal{\emph{et al}\onedot}

\begin{document}

\title{Generalizable Re-Identification from Videos with Cycle Association}
%
%
%
%

\author{Zhongdao Wang,
        Zhaopeng Dou, 
        Jingwei Zhang,
        Liang Zheng,~\IEEEmembership{Senior Memeber,~IEEE}, 
        Yifan Sun,
        Yali Li,~\IEEEmembership{Memeber,~IEEE}, 
        Shengjin Wang,~\IEEEmembership{Senior Memeber,~IEEE}
\IEEEcompsocitemizethanks{\IEEEcompsocthanksitem Z. Wang, Y. Li and S. Wang are with the Department
of Electronic Engineering, Tsinghua University, Bejing,
China, 100084.\protect\\
E-mail: wcd17@mails.tsinghua.edu.cn, \{liyali13,wgsgj\}@tsinghua.edu.cn
\IEEEcompsocthanksitem L. Zheng is with School of Computing, Australian National University, Canberra, ACT 2601, Australia.\protect\\
E-mail: liang.zheng@anu.edu.au 
\IEEEcompsocthanksitem J. Zhang is with Shanghai AI Lab, Shanghai, China.
\IEEEcompsocthanksitem Y. Sun is with Baidu Research, Beijing, China.
}
}

%
%

\markboth{Journal of \LaTeX\ Class Files,~Vol.~14, No.~8, August~2015}%
{Shell \MakeLowercase{\textit{et al.}}: Bare Advanced Demo of IEEEtran.cls for IEEE Computer Society Journals}
%



\IEEEtitleabstractindextext{%
\begin{abstract}

In this paper, we are interested in learning a generalizable person re-identification (re-ID) representation from unlabeled videos. 
Compared with 1) the popular unsupervised re-ID setting where the training and test sets are typically under the same domain, and 2) the popular domain generalization (DG) re-ID setting where the training samples are labeled, our novel scenario combines their key challenges: the training samples are unlabeled, and collected form various domains which do no align with the test domain. In other words, we aim to learn a  representation in an unsupervised manner and directly use the learned representation for re-ID in novel domains. 
To fulfill this goal, we make two main contributions:
First, we propose Cycle Association (CycAs), a scalable self-supervised learning method for re-ID with low training complexity; and second, we  construct a large-scale unlabeled re-ID dataset named LMP-video, tailored for the proposed method. 
Specifically, CycAs learns re-ID features by enforcing cycle consistency of instance association between temporally successive video frame pairs, and the training cost is merely linear to the data size, making large-scale training possible.
On the other hand, the LMP-video dataset is extremely large, containing 50 million unlabeled person images cropped from over 10K Youtube videos, therefore is sufficient to serve as fertile  soil for self-supervised learning.
Trained on LMP-video, we show that CycAs learns good generalization towards novel domains. The achieved results sometimes even outperform supervised domain generalizable models. 
\textbf{Remarkably, CycAs achieves 82.2\% Rank-1 on Market-1501 and  49.0\% Rank-1 on MSMT17 with zero human annotation}, surpassing state-of-the-art supervised DG re-ID methods by +0.2\% and +14.9\%, respectively. Moreover, we also demonstrate the superiority of CycAs under the canonical unsupervised re-ID (\emph{i.e.}, the same-domain setting) and the pretrain-and-finetune scenarios.

\end{abstract}

\begin{IEEEkeywords}
self-supervised Learning, person re-identification, cycle consistency
\end{IEEEkeywords}}

\maketitle

\IEEEdisplaynontitleabstractindextext

%
\IEEEpeerreviewmaketitle

\ifCLASSOPTIONcompsoc
\IEEEraisesectionheading{\section{Introduction}\label{sec:introduction}}
\else
\section{Introduction}
\label{sec:introduction}
\fi

\IEEEPARstart{W}{hile} progresses in metric learning have shown impressive improvements in person re-identification (re-ID), a great challenge, \ie, the poor generalization capability on unseen data domains, still remains to be a de facto drawback. 
Current methods have achieved an unprecedented retrieval accuracy in canonical datasets where the training and test data come from an identical domain, say $98\%$ Rank-1 on Market-1501~\cite{market}. However, when transferred to unseen domains, such representation significantly degenerates, with a considerable performance drop at about $20\%$ to $40\%$ in terms of Rank-1. 
This substantially limits the practical application of re-ID models as we definitely want a trained model to be deployed in various scenarios (domains), and even better, without adaptation.

\begin{figure*}[tbh]
    \centering
    \includegraphics[width=\linewidth]{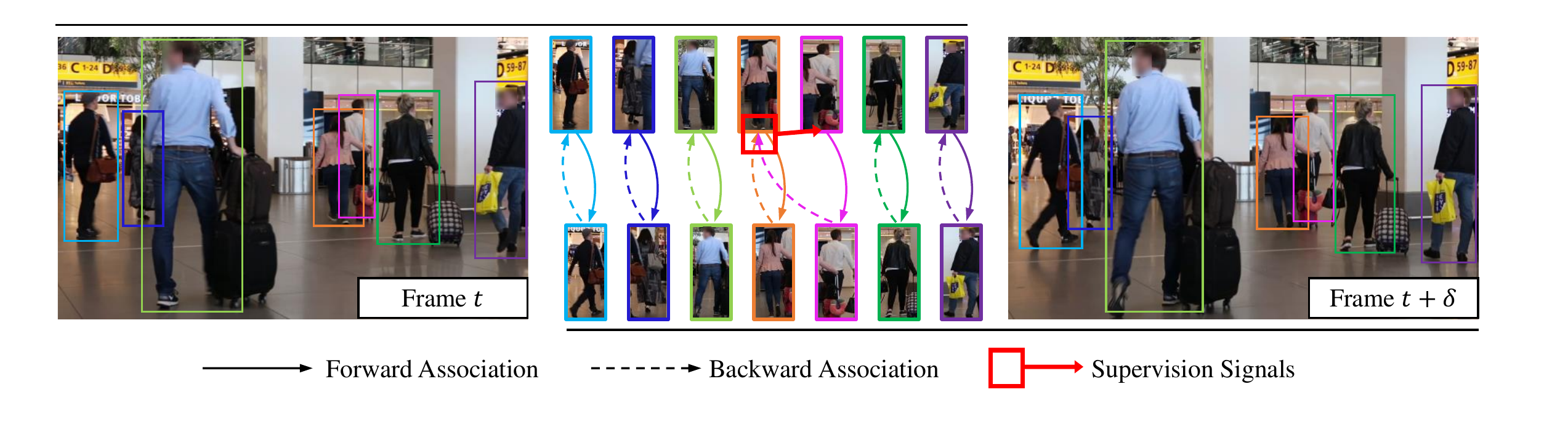}
    \caption{Core idea of the proposed Cycle Association (CycAs) pretext task. We perform forward and then backward association between two sets of person instances drwan from a pair of temporally consecutive video frames, and expect each instance to be associated to itself after the cycle. The inconsistency therefore could be used as ``free'' supervisional signals to learn a discriminative representation.}
    \label{fig:intro}
\end{figure*}

In this work, our ultimate goal is a universal re-ID representation that generalizes well across different test domains. We gain insights from the success of the face recognition community~\cite{arcface,ms1m,circle,frvt}, and factorize the key elements to achieving the above goal into two fold: 1) {Large-scale} training data collected from a {diverse distribution}, and 2) a {scalable} algorithm that can make full use of the massive data. We carefully design the two elements with a consideration of their close connection.

In terms of algorithm, we argue that unsupervised learning is promising and has several advantages over supervised learning.
The first is that the scalability of self-supervised methods has  been well-explored in recent literature. For instance, large language models~\cite{bert,gpt3} show a simple mask language modeling method scales well to massive data. Give the similar nature of representation learning, it is interesting to study how far self-supervised learning can go for re-ID.
More importantly, another consideration is the dispensation with manual annotations.
There is no doubt that the supervised methods (\eg, ArcFace~\cite{arcface}, Circle Loss~\cite{circle}) are scalable, but it is extremely difficult and costly to label a large-scale dataset for re-ID due to the quadratic complexity of annotation. With the help of self-supervised learning, the training data is much easier to be scaled up.

In terms of the training data, we argue two requirements should be met to ensure good  generalization. 
The first is apparently that we hope the amount of the training data to be as \emph{large} as possible.  Existing (labeled) re-ID datasets are typically with size $10^4$ to $10^5$ in terms of number of images. Practice from the face recognition community~\cite{ms1m,megaface} tells that this is at least two magnitudes smaller than we need for a generalizable representation. Although it becomes much easier to scale up the unlabled data, how to collect them still remains challenging if we pose a high requirement in amount, say $10^8$.
The second desired property is that the training data should be as \emph{diverse} as possible, so that a single domain hardly dominates. We notice that existing re-ID datasets are mostly collected from fixed surveillance cameras, therefore it is unavoidable that the low-level features of images bias to several fixed ``camera styles'', which in turn leads to poor generalization across domains. To tackle this tissue, domain diversity must be considered during the data collection process. 

As the first step towards our goal, we build a dataset named \emph{LMP-video}. 
LMP-video is a Large-scale, Multi-Person video dataset, composed of over 50M unlabeled person images in total.
Inspired by the LUP~\cite{LUP} dataset, we first crawl over 20K streetview videos from Youtube.
The videos are queried by keywords of 100 city names all around the world, ensuring sufficient diversity in data domains. For each video, we detect shot boundaries and split it into multiple continuous clips. Finally, we detect and crop all pedestrian in video clips and record the corresponding clip and frame index, as the final format of the training data. 

As a step beyond, we propose a novel self-supervised learning method to take full advantage of the LMP-video dataset.   The core idea is inspired by the multi-object tracking (MOT) task: given a video containing multiple pedestrian, tracking the persons forward and then backward along time should end up with the beginning instances. If not,  the errors  could serve as free supervisional signals for the model to learn a  discriminative appearance representation. We term this pretext task as Cycle Association (\emph{}{CycAs}), and show a brief illustration of the learning signal in \figref{fig:intro}.

\begin{table*}[tbh]
    \centering
    \caption{Comparison of existing re-ID experimental setups. }
    \begin{tabular}{l c c c l}
    \toprule
         \multirow{2}{*}{Setting} &  Train w/  & Adapt. w/  & Test  & Representative \\
         & ID label & ID label & domain & existing methods \\ 
         
         \midrule
         Supervised re-ID & {\Checkmark} & {No Adapt.} & {seen} & IDE~\cite{market}, PCB~\cite{pcb}, ArcFace~\cite{arcface},  CircleLoss~\cite{circle}, MGN~\cite{mgn} \\
         
         Unsupervised re-ID & {\XSolidBrush} & {No Adapt.}&  {seen} & PUL~\cite{pul}, BUC~\cite{buc}, UGA~\cite{uga}, SpCL~\cite{spcl},  \\
         
         UDA re-ID & {\Checkmark} & {\XSolidBrush}   &  {seen}  & SPGAN~\cite{dengUDA}, PTGAN~\cite{msmt}, MMT~\cite{mmt}, SpCL~\cite{spcl}\\
         
         Unsup. Pretrain$\rightarrow$ Sup. &{\XSolidBrush}   &{\Checkmark} &  {seen} & LUP~\cite{LUP}, ViT-pretrain~\cite{vitpretrain} \\
         
         Unsup. Pretrain$\rightarrow$ Unsup. & {\XSolidBrush}   &{\XSolidBrush}   &  {seen} & LUP~\cite{LUP}, ViT-pretrain~\cite{vitpretrain} \\

         DG re-ID & {\Checkmark} & {No Adapt.}& {unseen} & DIMN~\cite{DIMN}, SNR~\cite{SNR}. EasyReID~\cite{reidibn}, RaMoE~\cite{ramoe}\\
         
         \textbf{Unsupervised DG re-ID (this work)} & {\XSolidBrush}  &{No Adapt.} & {unseen} & CycAs (conference version of this work), TrackContrast~\cite{trackcontrast}  \\
         \bottomrule
    \end{tabular}
    
    \label{tab:category}
\end{table*}

Trained on the LMP-video dataset, the proposed CycAs representation shows strong generalization. 
We evaluate in the domain generalization (DG) re-ID setting, and show the unsupervised CycAs representation already performs favorably against state-of-the-art supervised methods.
Specifically, CycAs with ResNet-50 backbone achieves $80.3\%$ Rank-1 on Market-1501~\cite{market} and $43.9\%$ Rank-1 on MSMT17~\cite{msmt}. The Rank-1 score on the challenging MSMT17 outperforms the supervised counterpart trained with a joint set of multiple labeled datasets by $+9.8\%$. 
Furthermore, when switching the model architecture to a more data-hungry vision transformer, \eg, Swin-Transformer~\cite{swin}, we find the performance could be further improved, reaching $82.2\%$ Rank-1 on Market-1501 and $49.0\%$ on MSMT17, refreshing state-of-the-art in the DG re-ID setting.
We also evaluate our method on other experimental settings, such as unsupervised person re-ID, re-ID pre-training, and results show consistent superiority to previous methods. 
Most importantly, we show \emph{CycAs has good scalability to the size of training dataset}: its performance constantly improves when we scale up the training data, and we have not yet observe saturation at the current data size (50M images).
With increasingly growing data scale and capacity of neural architecture, we believe CycAs has a great potential to broaden the boundary of generalizable re-ID in the near future.

To summarize, our contributions are

\begin{itemize}
    \item We propose CycAs, a novel pre-text task for  self-supervised learning of person re-ID representations. The learned representation can be directly used without fine-tune or adaptation.
    \item We propose a large-scale, unlabeled, multi-person video dataset, LMP-video, tailored for the training of the proposed CycAs pre-text task. 
    \item Trained on LMP-video datset, the unsupervised CycAs representation achieves state-of-the-art performance on the  Domain Generalization (DG) re-ID task, outperforming previous \emph{supervised} representations.
    \item  we show CycAs has good scalability to the size of training dataset, and the scalability has not yet saturated at the current data scale.
\end{itemize}

\section{Related Work}

Person re-identification (re-ID) is an active topic of research in the past
decade,  fuelled by challenging benchmarks~\cite{market,duke,msmt,cuhk03}. 
Overall,  person re-ID can be seen as a specific case of the metric
learning~\cite{triplet,contrastive} problem, in which the goal is to learn a
metric space where instances of the same identity are grouped together and
instances from different identities are far apart. 
While metric learning methods do well in the re-ID task, a major issue still
remains unsolved, that so-learned representation presents rather unsatisfactory
generalization capability when used in unseen domains. To address this issue,
previous literature has fragmented the task in a multitude of different
experimental setups, which we detail as follows and summarize in \tabref{tab:category}.

\textbf{Supervised setting.} Supervised re-ID is the most well-explored
setting in which training and test data are drawn from an i.i.d. distribution,
and the domain generalization issue is simply not taken into account. In this
setting, previous works make efforts to adapt deep metric learning methods to
the data space of person images, by designing tailored model structures (such
as part models~\cite{pcb,mgn}) or loss functions~\cite{circle,bagoftricks}. 
Existing methods have achieve remarkable performance under this setting, \eg,
with a close-to-saturation Rank-1 CMC sore of over $98\%$ on standard benchmarks like Market-1501~\cite{market}.

\textbf{Unsupervised setting.} Similar to the supervised setting, most unsupervised 
re-ID methods also assume no domain gap exists between training and test data.
Specifically, a common experimental setting is to perform training and evaluation on
given train/test splits of an identical dataset. The only difference is that the
representation is learned in an unsupervised manner. Representative methods usually
resort to pseudo labels generated by performing clustering
algorithms~\cite{pul,buc,spcl} or by considering tracklet information~\cite{uga}. 
Given pseudo labels, the unsupervised setting could be solved in a manner similar to  
supervised methods. 

\textbf{Unsupervised Domain Adaptation (UDA) setting.} In this setting, 
it is assumed that a labeled \emph{source} dataset is given for training, and the
goal is to test on another \emph{target} dataset that drawn from a different
domain. The key problem here is how to adapt the representation to fit the
target domain. The adaptation is not blind, a set of \emph{unlabeled} data from the
target domain can be used as algorithm input.  Earlier methods usually opts to
transfer the distribution of the source domain data to match the target domain via
generative models such as GAN~\cite{dengUDA,msmt}, while more current progress shows
that clustering-based approaches~\cite{mmt,spcl} have superior performance. The main
drawback of the UDA setting is that the  representation is adapted to only a single
target domain, so that an extra adaptation is needed each time the target domain
changes.

\textbf{Unsupervised Pretraining setting.} 
Pre-training a representation on a large dataset is a prevalent way to improve
domain generalization capacity. Recent advances in general-purpose self-supervised
learning has shown a promising future of the pretrain$\rightarrow$finetune paradigm,
especially considering the eliminated requirement of manual data annotation. In the
re-ID task, there are several interesting recent efforts on unsupervised pretraining.
They run general-purpose self-supervised learning method, \eg,  MoCo~\cite{moco} and
DINO~\cite{dino}, on a large-scale unlabeled person dataset LUP~\cite{LUP}, and
explore the usage of vision transformers~\cite{vit,vitpretrain} in such cases. For
evaluation, a common practice is to use the learned representation for 
initialization, and finetune the model on target datasets in either a
supervised or an unsupervised manner. Trained on a large-scale dataset, the pretrain
model consistently improves upon the original ImageNet pretrain weights, refreshing
state-of-the-arts on the supervised and unsupervised re-ID setting. A primary
drawback of unsupervised pretraining models is that the pretrained representation
performs poorly if used for retrieval directly (without fine-tuning), say $2\%$ in terms of mAP in
Market-1501. Such poor performance can be mainly attributed to mismatch of  objectives  between the
pretraining pretext task (instance discrimination) and the re-ID task (identity
discrimination).

\textbf{Domain Generalization (DG) setting.} Among all, DG re-ID is the most
deployment-friendly setting, as it requires a representation, once trained, to be
generalizable across various unseen domains. Meanwhile, it is a most challenging
setting because any kind of domain adaptation is not allowed after training. In this
setting, supervised learning is feasible because it is assumed that labels of a
limited training set is available. Existing efforts mainly seek to design
generalizable model architecture~\cite{DIMN}, with especially consideration on the
normalization layers~\cite{SNR,reidibn} (\eg, BatchNorm~\cite{bn},
InstanceNorm~\cite{in}, IBN~\cite{ibn}, \etc). Besides these, RaMoE~\cite{ramoe}
shows effectiveness of Mixture of Experts (MoE) models in the DG setting. We believe
the DG setting is the most valuable one worthwhile exploring for real-world
applications in the current stage. However, a most critical meanwhile unsolved issue
is that the prohibitive cost of identity annotation hinders us from scaling up the
data size.

\textbf{Unsupervised Domain Generalization setting.}
In this setting, the test scheme is identical to the DG setting, but the training is
required to be \emph{unsupervised}. Obviously this suggests the setting is even more 
challenging than DG re-ID. Nevertheless, unsupervised learning brings an important
advantage, that it is easy to scale up the data size with negligible extra cost. With adequate
large-scale data and data-hungry algorithm, it is natural to expected unsupervised
models could even outperform supervised ones, which are trained with limited, relatively  small-scale data. This setting has not been well-explored in the re-ID community, to the best of knowledge, there are merely two related works. The first is the conference version of this work, and the second is TrackContrast~\cite{trackcontrast}. In the latter work, a large-scale training set comprising videos of 400+ hours is collected. The model is trained with a joint objective of tracklet classification and instance discrimination. 

  In this work, we mainly validate the effectiveness of CycAs on the most challenging \emph{unsupervised DG re-ID setting}. For the sake of fair comparison with existing methods, we also provide results in other settings.

\section{Self-supervised Cycle Association}
\label{sec:method}

\begin{figure*}[t]
    \centering
    \includegraphics[width=0.8\linewidth]{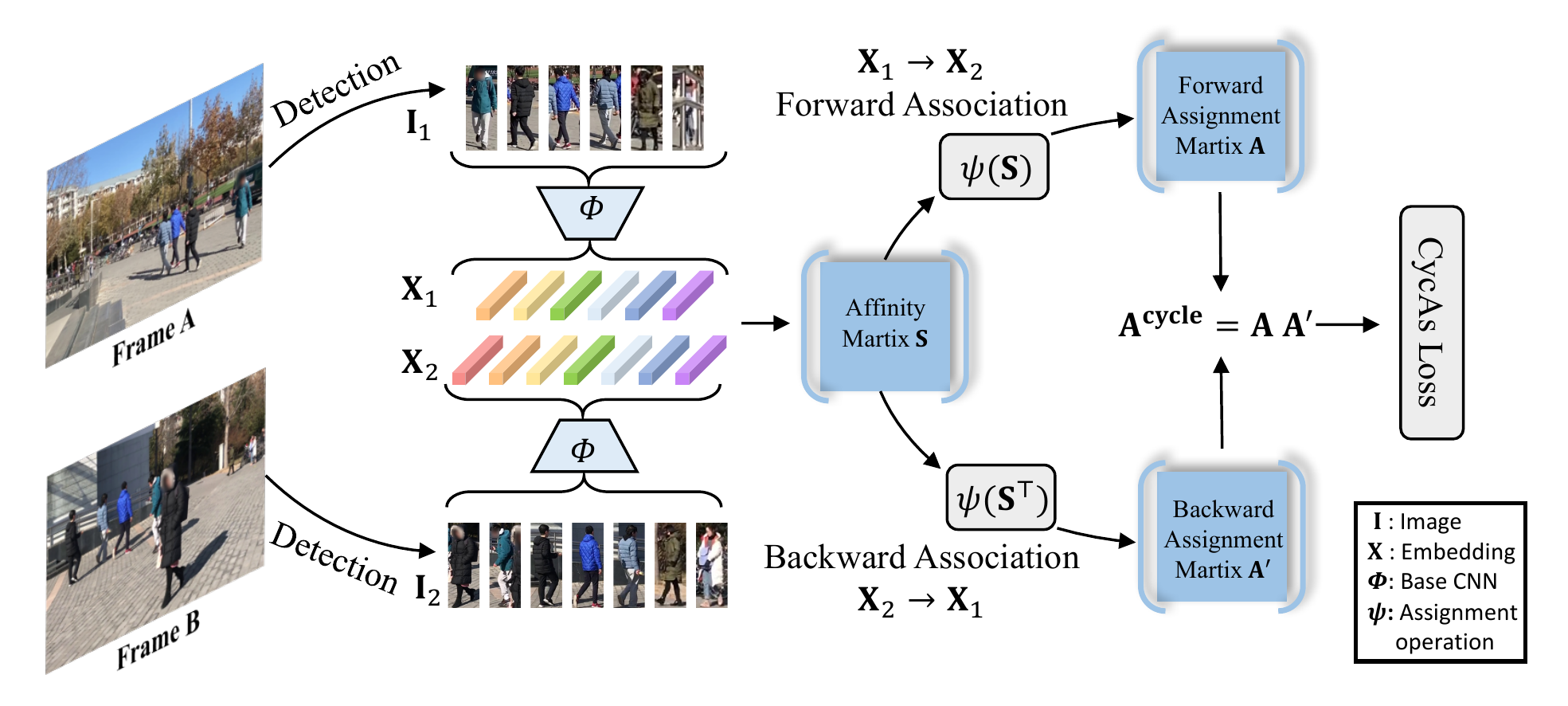}
    \caption {An overview of the proposed cycle association task. First, two sets of detected pedestrians are mapped to embedding vectors via function $\bm{\phi}$. Pairwise affinity matrix $\textbf{S}$ is computed from the two sets of embedding vectors, and then forward and backward assignment matrices $\textbf{A}$ and $\textbf{A'}$ are computed from $\textbf{S}$. Finally,difference between the cycle association matrix $\textbf{A}^\textrm{cycle} =\textbf{AA'}$ and the identity matrix is used as the supervision signal.}
    \label{fig:overview}
\end{figure*}

\subsection{Overview}
 We start with introducing the CycAs pretext task. 
 Our goal is to learn a discriminative pedestrian embedding function $\bm{\phi}$ by learning
 correspondences between two person images sets $\mathcal{I}_1$ and $\mathcal{I}_2$.
 Specifically, $\mathcal{I}_1$ and $\mathcal{I}_2$ are cropped from detected pedestrian
 bounding boxes in a pair of frames, and the frame pair is sampled within a short temporal interval from a video.
With proper selection of temporal interval, a reasonably high identity overlap between
$\mathcal{I}_1$ and $\mathcal{I}_2$ can be guaranteed. We define $\tau = \frac{\# \;
overlapped \; IDs}{max\{ \vert \mathcal{I}_1 \vert, \vert \mathcal{I}_2 \vert \}}$ as
the \emph{symmetry} between $\mathcal{I}_1$ and $\mathcal{I}_2$. For ease of
illustration, let us begin with the absolute symmetric case $\tau=1$, \ie, for any
instance in $\mathcal{I}_1$ there always exists a correspondence in $\mathcal{I}_2$, 
and vice versa. Then we introduce how we deal with the  asymmetry case ($\tau<1$) in
Section~\ref{sec:method:asymmetry}. An overview of the training procedure is presented
in Fig.~\ref{fig:overview}.

\subsection{Association between Symmetric Pairs}
Consider all bounding boxes in $\mathcal{I}_1 \cup \mathcal{I}_2$ forming a minibatch. Under
the assumption that $\tau = 1$, \ie, absolute symmetry, we have $\vert \mathcal{I}_1
\vert = \vert \mathcal{I}_2 \vert = K$. The images are mapped into an embedding space
by a function $\bm{\phi}$, such that  $\mathbf{X}_1 = \left[ \mathbf{x}_1^1, \mathbf{x}_1^2,\cdots,\mathbf{x}_1^n\right]$
and $\mathbf{X}_2 = \left[ \mathbf{x}_2^1, \mathbf{x}_2^2,\cdots,\mathbf{x}_2^n\right] \in \mathbb{R}^{d\times n}$, where the embedding matrices are composed of $n$ embedding vectors  of dimension $d$. 
All the embedding vectors are $\ell_2$-normalized. To capture similarity between instances, we compute an affinity matrix between all instances in $\mathbf{X}_1 $ and $ \mathbf{X}_2$ by pairwise cosine similarity, 
\begin{equation}
\label{eq:sim}
     \mathbf{S} = \mathbf{X}_1^\top \mathbf{X}_2 \in \mathbb{R}^{n\times n}.
\end{equation}

We take affinity matrix $\mathbf{S}$ as the input to perform \emph{data association}. This process aims to predict correspondences in $\mathcal{I}_2$ for each instance in $\mathcal{I}_1$. Formally, the goal is to obtain an assignment matrix 
$
    \mathbf{A} = \bm{\psi} (\mathbf{S}) \in \{0,1\}^{n\times n},
$
where $1$ indicates correspondence, and $0$ non-correspondence. We are especially inspired by the multi-object-tracking (MOT) task, in which data association has a critical role. In MOT, a common practice is to model association as a linear assignment problem, and the solution $\bm{\psi}$ can be given by the Hungarian algorithm (examples can be found in many MOT algorithms~\cite{deepsort,poi,jde}). 
The association could be performed backward, \ie, from $\mathcal{I}_2$ to $\mathcal{I}_1$, and we denote the backward association matrix as $\mathbf{A}'$.

Suppose the embedding function is perfect, \ie, the cosine similarity between vectors of the same identity equals~$1$, and the cosine similarity between vectors from different identities equals~$-1$. The Hungarian algorithm would output the optimal assignment $\mathbf{A}^{*}=\frac{\mathbf{S} + 1}{2}$ for the forward association process $\mathbf{X}_1 \rightarrow \mathbf{X}_2$. The backward association process $\mathbf{X}_2 \rightarrow \mathbf{X}_1$ is similar, and the optimal assignment matrix $\mathbf{A}'^{*} =  \frac{\mathbf{S}^\top + 1}{2}$. 
A cycle association pass is then defined as a forward plus a backward association pass,
\begin{equation}
\label{eq:cycle}
    \mathbf{A}^\textrm{cycle} = \mathbf{A} \mathbf{A}'.
\end{equation}

If $\mathbf{A}=\mathbf{A}^{*}$ and $\mathbf{A}'=\mathbf{A}'^{*}$, the forward and back association processes are optimal, allowing an instance to  correctly match its appearance in the other frame and then correctly match back to itself. In other words, under optimal association,  an instance will be associated to itself after the cycle. 
Equivalently, the cycle association matrix underpinning perfect association
$\mathbf{A}^\textrm{cycle}$ should equal the identity matrix $\mathbf{I}$.
Therefore, the difference between $\mathbf{A}^\textrm{cycle}$ and $\mathbf{I}$ can be used as signals to implicitly supervise the model to learn correspondences between $\mathbf{X}_1 $ and $\mathbf{X}_2$.

In order to perform end-to-end training to learn a meaningful representation, we need the cycle association to be differentiable. 
However, the assignment operation $\bm{\psi}$ (Hungarian algorithm) is not differentiable. This drives us to design a differentiable $\bm{\psi}$. We notice that if the one-to-one correspondence constraint in linear assignment is removed, $\bm{\psi}$ can be approximated by the row-wise \texttt{argmax} function. Considering \texttt{argmax} is not differentiable either, we further soften this operation by the row-wise softmax function. Now, the assignment matrix is computed as,
\begin{equation}
\label{eq:softmax}
    A_{i,j} = \psi_{i,j} (\mathbf{S}) = \frac{e^{T{S}_{i,j}}}{\sum_{j'}^k  e^{T{{S}_{i, j'}}}},
\end{equation}
where ${A}_{i,j}$ is the element of $\mathbf{A}$ in the $i$-th row and the $j$-th colomn, and $T$ is the temperature of the softmax operation.

Combing \equref{eq:sim}, ~\equref{eq:softmax} and \equref{eq:cycle}, a cycle association matrix $\mathbf{A}^\textrm{cycle}$ can be computed with all operations therein being differentiable. Finally, the loss function is defined as the mean $\ell_1$ error between $\mathbf{A}^\textrm{cycle}$ and $\mathbf{I}$,
\begin{equation}
\label{eq:symmetricloss}
    \mathcal{L}_\textrm{symmetric} = \frac{1}{n^2} \Vert \mathbf{A}^\textrm{cycle} - \mathbf{I} \Vert_1.
\end{equation}

\begin{figure}[t]
    \centering
    \includegraphics[width=\linewidth]{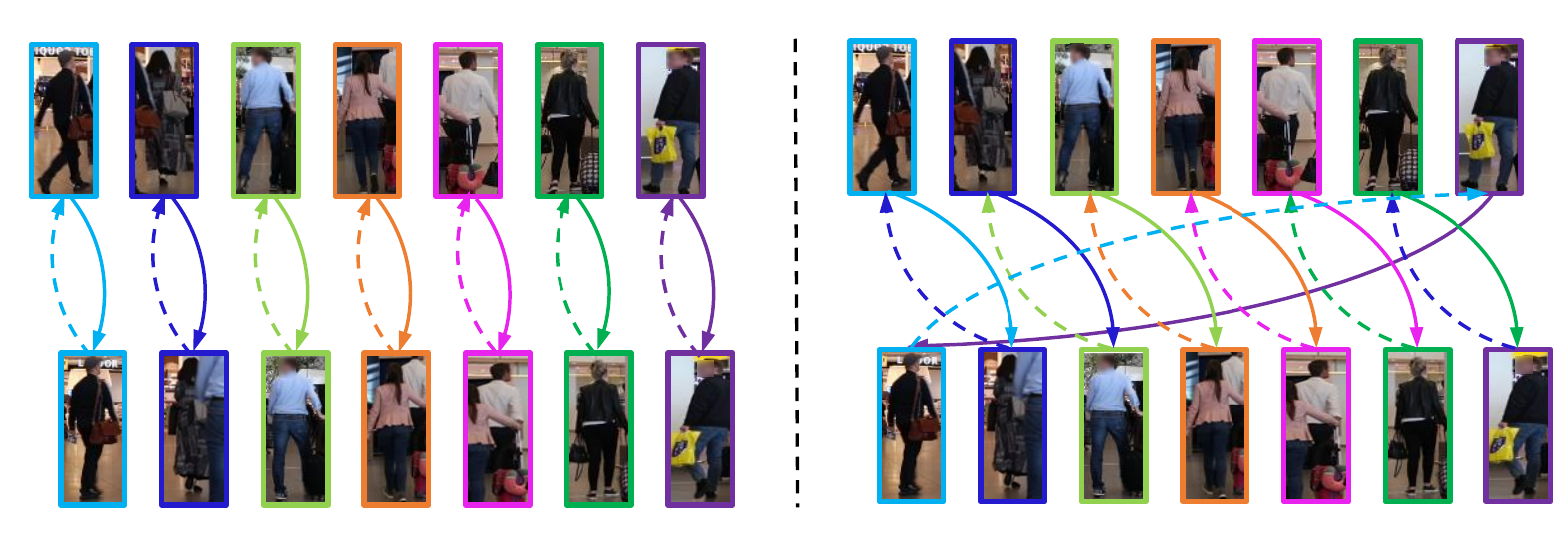}
    \caption{A non-trivial solution (\textbf{Left}) \vs a trivial solution (\textbf{Right}) of Eq. \ref{eq:symmetricloss}. Solid and dotted lines means forward and backward correspondences, respectively. We reasonably speculate the trivial solutions are very unlikely to be learned, which is confirmed in our experiment.}
    \label{fig:trivial}
\end{figure}

\textbf{Discussion.} Theoretically, {cycle consistency} is a \emph{necessary} but not \emph{sufficient} requirement for discriminative embeddings. 
In Fig.~\ref{fig:trivial} right, we present a trivial solution: what the embedding model has learned is matching an identity to the one with the exact next index, and the one with the last index matching to the first. Such a trivial solution requires the model to build correlations between completely random identity pairs, which share very limited, if any, similar visual patterns. Therefore, we reasonably speculate that by optimizing the cycle association loss, it is very unlikely for the model to converge to such trivial solutions and that it is much easier to converge to non-trivial solutions (Fig.~\ref{fig:trivial} left). This conjecture is empirically confirmed in experiments: the converged models always give discriminative embeddings, and trivial solutions  never emerge. 

\subsection{Relaxation on Asymmetric Pairs for Association}
\label{sec:method:asymmetry}

In practice, asymmetric always arises for a few reasons. 
First, pedestrians always enter and leave the camera field of view. 
Second, real-world videos usually contain persons with high-velocity motions, severe occlusions, and low-quality, so detectors may fail sometimes.
Although we can control the sampling interval between frame pairs to be short and ensuring high symmetry, the appearance changes of persons would be compromised accordingly. 
Sufficient appearance changes are of great value for learning a discriminative representation, therefore we must use a relatively long sampling interval. 
This means it is important to address the asymmetry issue, and  we make the following efforts to reduce its negative impact.
 
First, consider embedding matrices $\mathbf{X}_1$ and $\mathbf{X}_2$ with the number of embedding vectors being $n_1$ and $n_2$, respectively. According to \equref{eq:cycle}, the resulting $\mathbf{A}^\textrm{cycle}$ is of size $n_1 \times n_1$. 
Further, we know that $\texttt{rank}(\mathbf{A}^\textrm{cycle})\le \min(n_1, n_2)$. 
If $n_1 > n_2$, there will exist at least $n_1 - n_2$ instances that cannot be associated back to themselves. This will introduce ambiguity. In this case, we swap $\mathbf{X}_1$ and $\mathbf{X}_2$ in such cases to ensure $n_1 \le n_2$ always holds.

Second, we improve the learning objective. In the asymmetric scenario, the loss function $\mathcal{L}_\textrm{symmetric}$ is sub-optimal, because some instances may lose correspondences in cycle association, and thus the corresponding diagonal elements in $\mathbf{A}^\textrm{cycle}$ should not equal $1$. Simply changing the supervision of these lost instances from $1$ to $0$ is not feasible, because there are no annotations and we do not know which instances are lost. 
To address this, we relax the learning objective. More specifically,
we require a diagonal element ${A}^{\textrm{cycle}}_{i,i}$ to be greater than all the other elements along the same row and column, by a given margin $m$. The loss function is formulated as,

\begin{equation}
\label{eq:asymmetricloss}
\begin{aligned}
     \mathcal{L}_\textrm{asymmetric} = & \frac{1}{n_1} 
    \sum_{i=1}^{n_1} 
    \left(\max_{j\ne i} {A}^{\textrm{cycle}}_{i,j} - 
    {A}^{\textrm{cycle}}_{i,i} + m\right)_+ \\
    + 
    & \frac{1}{n_1}\sum_{i=1}^{n_1} 
    \left(
    \max_{k\ne i} {A}^{\textrm{cycle}}_{k,i} - {A}^{\textrm{cycle}}_{i,i} + m \right)_+,
\end{aligned}
\end{equation}
 Margin $m$ is a hyper-parameter ranging in $(0,1)$ with smaller values indicating softer constrains. We set $m=0.5$ in all the experiment if not specified. 

The relaxation of the loss function benefits learning in both the asymmetric and symmetric cases. In the experiment, we use $\mathcal{L}_\textrm{asymmetric}$ by default.

\subsection{Adapt Softmax Temperature to Varying Sizes}
\label{sec:method:temp} 
The Softmax operation, as we observe, has different levels of softening ability on inputs with different sizes.
Compared to a short vector, the maximum value in a longer vector is less highlighted after the Softmax function. To alleviate this problem and stabilize training, we let the Softmax temperature be adaptive to the varying input size, so that 
the maximum values in them are equally highlighted.

To fulfill this goal, we set the temperature $T = \frac{1}{\epsilon} \log \left[ \frac{\delta (K-1) + 1}{1-\delta} \right]$, where $\epsilon$ and $\delta$ are two hyper-parameters ranging from 0 to 1. In fact, our preliminary experiment indicates the only hyper-parameter that matters is $\epsilon$,and that $\delta$ can be simply set to $0.5$. This leads to the final form $T = \frac{1}{\epsilon} \log (K+1)$. Detailed derivation and discussion can be found in appendix.

\subsection{Negative Mining with Cross-batch Memory}

In our training data sampling strategy, a large chance is ensured that an instance could find its positive sample in a minibatch, but the number of negative samples are rather limited. In a minibatch, all samples come from a single video , and we can only contrast an instance to a fixed set of negative samples co-appearing in this video. The limited negative samples compromise representation learning to some extent, and we propose the following improvements.

First, during training, we sample multiple frame pairs instead of a single one to form the minibatch. From each frame pair indexed with $i$, we detect and crop all persons, forming two image sets $\mathcal{I}_1^{i}$ and $\mathcal{I}_2^{i}$. Then we merge corresponding image sets from multiple videos, \ie,  $\mathcal{I}_1 = \bigcup_{i} \mathcal{I}_1^{i}$ and  $\mathcal{I}_2 = \bigcup_{i} \mathcal{I}_2^{i}$. Sampling multiple frame pairs in a minibatch not only brings more negative samples, but also  boosts training efficiency significantly. 

However, the batch size cannot grow unlimitedly due to limited GPU memory. Inspired by Wang \etal~\cite{xbm}, we introduce a cross-batch memory (XBM) to store more negative samples. XBM is implemented with a queue of length $M$, and each element in the queue is a feature vector computed from the past iterations.  In each training iteration, the memory is updated in a dequeue-enqueue manner. XBM is memory efficient because the feature vector it stores is compact and does not require gradients. Specifically, a large XBM with $M=65536$ and feature dimension $d=512$ costs only 128MB extra GPU memory.

In contrast to the original implementation, we additionally store a \emph{video ID} along with each feature vector in the memory. For each minibatch feature $\mathbf{x}_i$ and its video ID $v_i$, we regard features with different video ID as negative samples. Finally, we minimize similarities between batch samples and all the negative samples in the memory by:

\begin{equation}
    \mathcal{L}_\mathrm{XBM} = \frac{1}{NM}\sum_{i=1}^{N}\sum_{j=1}^{M} \mathds{1} (v_i \ne v_j) \mathrm{Softplus} ( \mathbf{x}_i^\top \mathbf{z}_j),
\end{equation}
where $\mathbf{z}_j$ represent the $j$-th feature in the XBM, $\mathds{1}(\cdot)$ is the indicator function, and $\mathrm{Softplus}(x) = \log(1+e^x)$.

In practice, we find \emph{hard negative samples} more valuable in the training process.  Instead of considering all negative samples in the memory, we select the top $k$ nearest samples \wrt a batch sample and discard all the others when calculating the loss. We set $k=10$ unless specified.

\section{The LMP-video Dataset}
\label{sec:dataset}

Training data is a critical ingredient to improve generalization ability. 
In our context, we expect a dataset to have the following properties.
First and the most important, the dataset should be large and diverse in types of appearance. 
Based on this point, we find unlabeled videos could be fertile soil for our purpose.
Second, we hope videos contain multiple persons. This means we can sample fewer frame pairs for a mini-batch with fixed size, thus improving the sampling efficiency.  
Third, we want to have possibly high symmetry $\tau$ between the sampled frame pair, which reduces noise in training. 
Last, we expect relatively large appearance changes of a person between the a frame pair.

\begin{figure}[t]
    \centering
    \includegraphics[width=\linewidth]{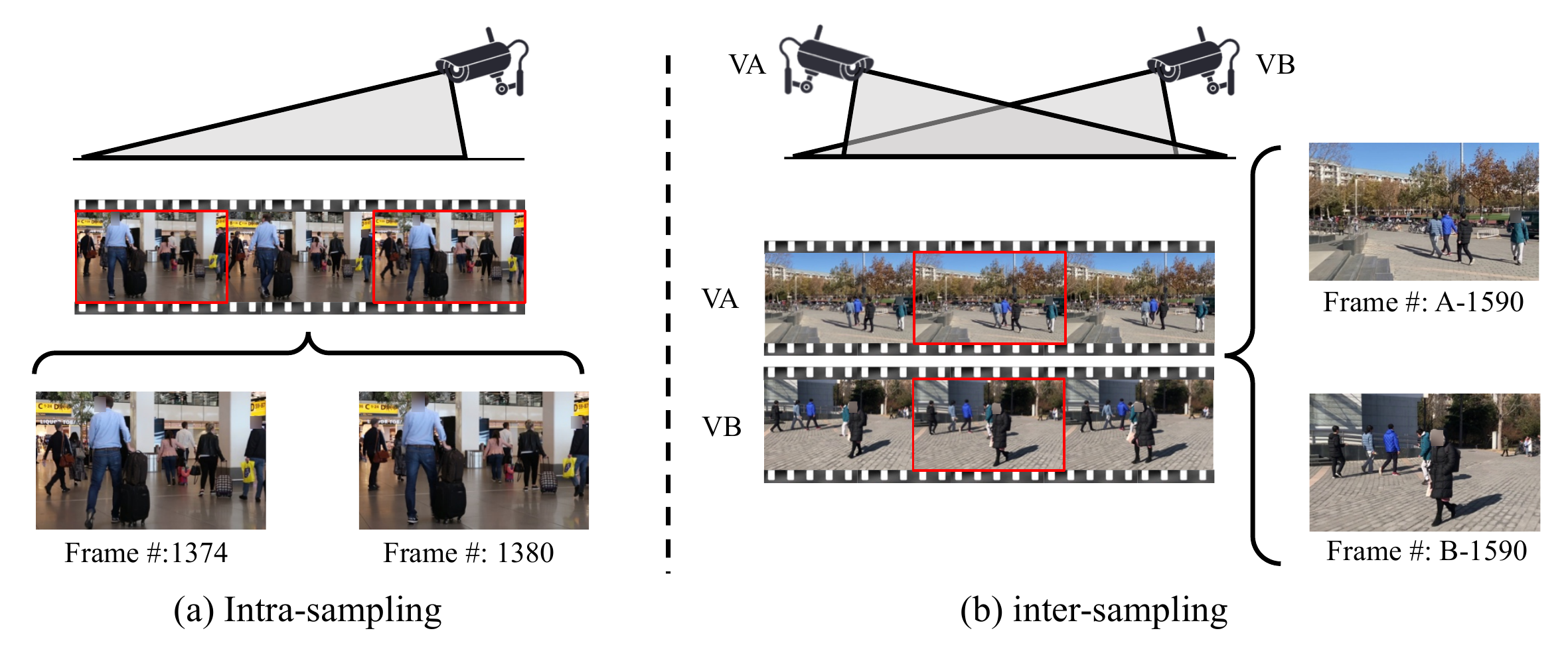}
    \caption{Illustration on the two data sources \textbf{(a) Intra-sampling:} frame pairs are drawn from a single video within a short temporal interval.  \textbf{(b) Inter-sampling:} frame pairs are drawn from cameras capturing an overlapped field of view, at the same time.  }
    \label{fig:sampling}
\end{figure}

The latter two properties contradict with each other, which poses a interesting challenge. To achieve a trade-off effect, 
we propose that the collected data allow for the following two data sampling strategies (see \figref{fig:sampling}).  
\begin{itemize}
    \item \emph{Intra-video pair sampling.}  A frame pair are sampled from the same video within a short temporal interval.
    \item \emph{Inter-video pair sampling.} A frame pair is sampled at the same timestamp from two synchronized cameras that capture an overlapped field of view.
\end{itemize}

Data allowing for intra-video pairs are easy to collect and have a relatively high symmetry, but lack appearance changes. The inter-video pairs have large appearance changes, but the data symmetry is poor, and one must collect \emph{paired} videos from scratch because there is no off-the-shelf resources on the Internet. To investigate how to take full advantage of existing data resources, we collect a large amount of data of both the two types, forming a \textbf{L}arge-Scale \textbf{M}ulti-\textbf{P}erson video dataset (\textbf{LMP-video}). Data collecting and processing procedures are detailed in the following subsections, for the two types of data respectively. 

\begin{figure}
    \centering
    \includegraphics[width=\linewidth]{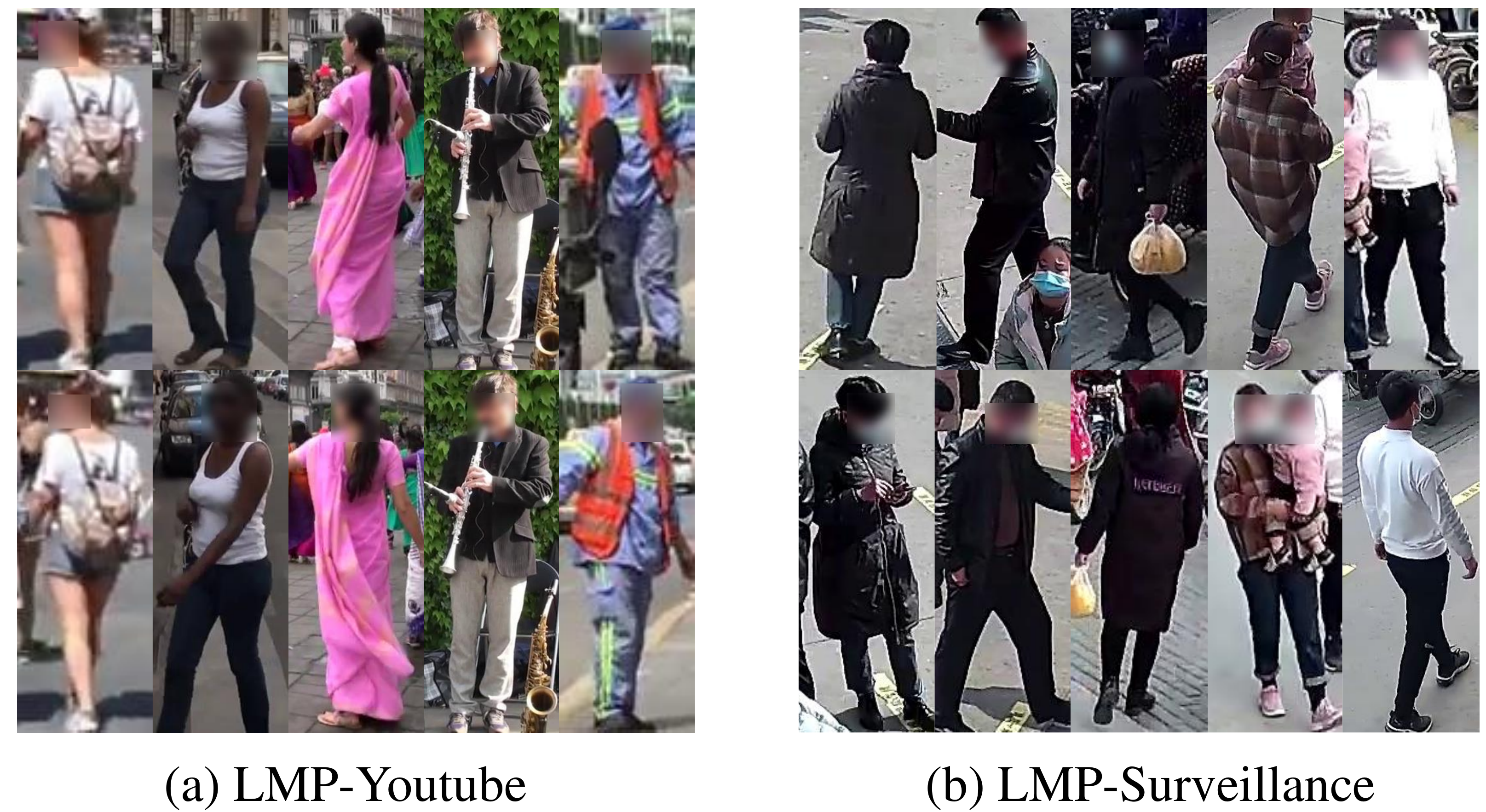}
    \caption{Examples of the LMP-video dataset. (a) the LMP-Youtube subset is more diverse in terms of person appearance but videos are captured from a single view. (b) the LMP-surveillance subset contain paired videos captured from two views but relatively lacks appearance diversity.}
    \label{fig:LMP}
\end{figure}

\subsection{Intra-video Sampled Data}
\label{sec:dataset:intra}
We collect intra-video data are from Internet videos. Specifically, we crawled over 10K videos from Youtube. The video list is originally provided by the LUP~\cite{LUP} dataset,  structured by querying the search engine with strings like \texttt{cityname + streeview}. The \texttt{cityname} can be that of 100 big cities all around the world, and the number of videos is balanced for each city.
For fast development iteration, we only select the 10K videos out of the whole 70K in the list. 
We find most Internet videos are edited by splicing multiple shots, so that the video is not continuous in terms of contents right before and after the shot changes. This may introduce wrong frame pairs in data sampling. Therefore, we cut raw videos into separate clips with the PySceneDetect\footnote{\url{http://scenedetect.com/}} toolkit, eliminating shot changes inside each clip. 
Next, in order to ensure videos contain multiple persons, videos with too few persons are removed. We run a person detector JDE~\cite{jde} on 5 uniformly sampled  frames of a given video, and compute the average number of detected persons. If the average number is less than 2, the video is removed.  
Finally, for the remaining videos, we detect and crop person images with the JDE detector. Detection is performed every 7 frames to reduce the temporal redundancy of videos. 
The resulting sub-dataset contains 74K short video clips with a total length of 220 hours,  5.4M frames, and 47.8M cropped person images. We denote this sub-dataset as \emph{LMP-youtube} in the following sections. \figref{fig:LMP}a shows several examples.

\subsection{Inter-video Sampled Data}
\label{sec:dataset:inter}
It is difficult to collect a large amount of paired videos from the Internet for inter-video sampling (\figref{fig:sampling}b) because of limited available resources. 
Instead, we have to collect such videos in real-world.  Specifically, we deploy two cameras at a gateway of a building, make the two cameras capture an overlapped field of view from different perspectives, and record videos when multiple pedestrian walk through. The videos are processed in a similar way as above (but no need of scene detecting). Finally, we obtain a sub-dataset containing 25 long video pairs with a total length of 50 hours,  62K video frames and 3.0M cropped person images. We denote this sub-dataset as \emph{LMP-surveillance} in the following sections. Note all the videos are captured in a \emph{single} scene in LMP-surveillance, thus lack domain diversity.  \figref{fig:LMP}b shows several examples.

\subsection{Take-away Messages}
\label{sec:dataset:summary}
Below we summarize several take-away-message on the LUP-video dataset that we discover in experiments.
\begin{itemize}
    \item \emph{Both the two data sources help.} We find training on both the two subsets leads to decent representation, and using the combination of the two yields the best.
    \item \emph{Scaling up Youtube data improves generalization.} When we keep scaling up the size of LMP-youtube, cross-domain re-ID accuracy constantly improves. We argue a critical factor behind such scalability is the domain diversity in Youtube videos.
    \item \emph{Scaling up surveillance data does not improve  generalization. } Similarly, we argue the main reason is the lack of domain diversity in surveillance videos.
\end{itemize}

Although we believe that collecting a domain-diverse surveillance data is still possible and would definitely help, it is obvious that a more cost-efficient option is to scale up Youtube data. Therefore, the final LMP-video dataset is composed of a very large LMP-youtube subset and a tiny LMP-surveillance subset. 
Empirical results and further discussions on the dataset will be described in \secref{sec:exp:dataset_scalability}.

\section{Experimental Results}
\label{sec:exp}
Below, we evaluate CycAs in a wide range of experimental settings. 
\secref{sec:exp:dg} presents the key results of this work where we evaluate in the unsupervised Domain Generalizable re-ID setting.
In \secref{sec:exp:unsup}, we evaluate in the unsupervised re-ID setting where domain gaps do not exist.
In \secref{sec:exp:pretrain}, we show CycAs can also serve as pretrain models.
Finally, in \secref{sec:exp:ablation}, we study key factors that affect the performance of CycAs.

\subsection{Unsupervised Domain Generalizable re-ID}
\label{sec:exp:dg}
\subsubsection{Training Details}
\label{sec:exp:dg:traindetail}
We employ ResNet-50 as the default architecture if not specified. Inspired by~\cite{ibn} and~\cite{reidibn}, in order to prevent the model from being biased to the training domain, we 1) append two additional Instance Normalization (IN) ~\cite{in} layers at the end of the first two stages, and 2) freeze parameters in all Batch Normalization (BN) layers during training. In one minibatch, we sample cropped person images from 16 frame pairs. The batch size, in terms of image number, therefore is varying, and we truncate it if it exceeds 80 (Sizes of image sets $\mathcal{I}_1$ and $\mathcal{I}_1$ truncated at 40 respectively), to prevent a minibatch grows beyond the fixed GPU memory limit.
We adopt distributed training on 8$\times$ NVIDIA A100 GPU, so the maximum total batch size in terms of image numbers is $640=80\times 8$.
We define a training epoch as when all videos in the dataset are sampled for 16 times. The model is optimized with the AdamW~\cite{adamw} optimizer for 50 epochs in total. In the first 45 epochs the model is trained on LMP-youtube data with Cosine Annealing learning rate~\cite{cosinelr} initialized with $1e^{-4}$. In the last 5 epochs we train on both LMP-youtube and LMP-surveillance data with  a fixed learning rate $2.5e^{-6}$.  Training takes 20 hours.

\begin{table*}[]
    \scriptsize
    \centering
    \caption{Comparison to existing domain generalizable re-ID methods. The superscript $^\dag$ indicates unsupervised methods, otherwise supervised. For supervised methods, Protocol-1 requires leave-one-out evaluation, \ie train on three of the four datasets (Market~\cite{market}, Duke~\cite{duke}, CUHK03~\cite{cuhk03} and MSMT17~\cite{msmt}) and test on the ramaining one. In Protocol-2, the training set is the joint set of Market+Duke+CUHK02+CUHK03+CUHK-SYSU~\cite{sysu}, and the test sets are four small re-ID datasets.
    Sup. indicates supervised or unsupervised method. 
    Unsupervised methods employ self-collected training data and are evaluated on test sets in each protocol, respectively.}
    \begin{tabular}{lccccccccc cccccccc}
    \toprule
        \multirow{3}{*}{Method} & \multirow{3}{*}{Sup.} & \multicolumn{8}{c}{Protocol-1}  & \multicolumn{8}{c}{Protocol-2}  \\
          \cmidrule(lr){3-10}
          \cmidrule(lr){11-18}
          &   &\multicolumn{2}{c}{Market~\cite{market}} &  \multicolumn{2}{c}{Duke~\cite{duke}} &  \multicolumn{2}{c}{CUHK03~\cite{cuhk03}} &  \multicolumn{2}{c}{MSMT17~\cite{msmt}}  
          &  \multicolumn{2}{c}{PRID~\cite{prid}} &  \multicolumn{2}{c}{GRID~\cite{grid}} &  \multicolumn{2}{c}{VIPeR~\cite{viper}} &  \multicolumn{2}{c}{iLIDS~\cite{ilids}}  
          \\ 
          & & mAP & R-1 & mAP & R-1 & mAP & R-1 & mAP & R-1  & mAP & R-1 & mAP & R-1 & mAP & R-1 & mAP & R-1 
          \\
          \midrule
          Agg\_Align~\cite{align}&	\Checkmark &	-	& - & - &	- & 	-	& - & - &	-& 25.5&	17.2&	24.7&	15.9&	52.9&	42.8&	74.7&	63.8\\
          CrossGrad~\cite{crossgrad}&	\Checkmark&	-	& - & - &	- & 	-	& - & - &	-& 28.2	&18.8&	16.0&	9.0&	30.4&	20.9&	61.3&	49.7\\
          PPA~\cite{PPA}&	\Checkmark&	-	& - & - &	- & 	-	& - & - &	-& 45.3&	31.9&	38.0&	26.9&	54.5&	45.1&	72.7&	64.5\\
          DIMN~\cite{DIMN}&	\Checkmark&	-	& - & - &	- & 	-	& - & - &	-& 52.0&	39.2&	41.1&	29.3&	60.1&	51.2&	78.4&	70.2\\
          SNR~\cite{SNR}&	\Checkmark&	-	& - & - &	- & 	-	& - & - &	-& 66.5&	52.1&	47.7&	40.2&	61.3&	52.9&	89.9&	84.1 \\
          M$^3$L\cite{m3l}&	\Checkmark & 50.2 &	75.9 &	51.1 &	\underline{69.2} &	32.1 &	{33.1} &	12.9 &	33.0 &	-	& - & - &	- & 	-	& - & - &	- \\
          DML~\cite{ramoe}&	\Checkmark& 49.9&	75.4&	49.4&	65.8&	{32.6}&	32.9&	9.9&	24.5&	60.4&	47.3&	49.0&	39.4&	58.0&	49.2&	84.0&	77.3 \\
          RaMoE~\cite{ramoe}&	\Checkmark&	{56.5}&	\underline{82.0}&	\textbf{56.9}&	\textbf{73.6}&	\underline{35.5}&	\textbf{36.6}&	{13.5}&	{34.1}&	{67.3}&	{57.7}&	{54.2}&	{46.8}&	{64.6}&	{56.6}&	{90.2}&	{85.0 }\\
          TrackContrast$^\dag$~\cite{trackcontrast}&	\XSolidBrush & 36.2	 & 72.7 & 	31.2&	51.7& 	-	& - & - &	- & 	-	& - & - &	- & 	-	& - & - &	-\\
          \textbf{CycAs\_R50}$^\dag$&	\XSolidBrush &  \underline{57.5}&	{80.3}&	{51.8}&	65.8&	26.5&	25.8&	\underline{20.2}&	\underline{43.9} & \underline{67.7}&	\underline{58.8}&	\underline{62.3}&	\underline{52.5}&	\underline{66.0}&	\underline{57.3}&	\underline{90.4}&	\underline{85.2} \\
          \textbf{CycAs\_Swin}$^\dag$ &	\XSolidBrush & \textbf{60.4}&	\textbf{82.2}&	\underline{56.2}&	{69.1}&	\textbf{37.1}&	\underline{36.3}&\textbf{24.1}&	\textbf{49.0} &\textbf{79.2}&\textbf{71.5}&	\textbf{66.4}&	\textbf{55.8}&	\textbf{68.8}&	\textbf{60.2}&	\textbf{91.3}&	\textbf{87.4}\\

          \bottomrule
         
    \end{tabular}
    
    \label{tab:exp:dg}
\end{table*}

\subsubsection{Evaluation Protocols}
There are few existing works that address domain generalizable (DG) re-ID in an unsupervised manner, therefore we mainly compare with supervised DG re-ID methods. In these works, results are reported using two prevalent evaluation protocols. Protocol-1 requests \emph{leave-one-out}  training and testing on four largest publicly available  datasets, \ie, Market-1501~\cite{market}, DukeMTMC-ReID~\cite{duke}, CUHK03~\cite{cuhk03}, and MSMT17~\cite{msmt}.  For instance, when testing on Market-1501, the joint set of the other three datasets are used for training.  
Protocol-2 instead requests evaluation on four small-scale datasets, \ie, PRID~\cite{prid}, GRID~\cite{grid}, VIPeR~\cite{viper}, and iLIDS~\cite{ilids}, while the training set is a  combination of several other non-overlapping datasets,  Market-1501, DukeMTMC-ReID, CUHK02~\cite{cuhk03}, CUHK03~\cite{cuhk03}, and CUHK-SYSU~\cite{sysu}. 
For protocol-2, We  report average results of 10 trials, in each of which the queries and the gallery are randomly selected strictly following the evaluation setting described in DIMN~\cite{DIMN}.  
Note that unsupervised methods do not need to follow the two protocols in terms of training. They employ large-scale self-collected data for training, and directly test on the given test sets in each protocol. 

\subsubsection{Results}
Comparisons are presented in \tabref{tab:exp:dg}. First, considering all existing methods use the ResNet-50 architecture, we compare a ResNet-50 based model (denoted as CycAs\_R50) with them.  A noticeable baseline to be compared with is deep metric learning (denoted as DML) presented in RaMoE~\cite{ramoe}.  In this baseline, supervised training is applied specifically in form of combining three loss functions: softmax cross-entropy loss, triplet loss~\cite{triplet}, and center loss~\cite{centerloss}. We observe results of the DML baseline quite competitive, already outperforming several former methods tailored for the DG re-ID task, such as M$^3$L~\cite{m3l} and DIMN~\cite{DIMN}. Being unsupervised, CycAs\_R50  outperforms the strong DML baseline in seven out of the eight test sets, by significant margins ranging from $+2.4\%$ mAP to $+13.3\%$ mAP. Another method to be compared with is the supervised state-of-the-art RaMoE~\cite{ramoe}. Compared with RaMoE, CycAs\_R50 is also superior in most datasets (six out of eight) in terms of mAP. In the largest test set MSMT17, CycAs even outperforms RaMoE by $6.7\%$ mAP.  The above comparisons suggest that, by leveraging large-scale unlabeled data, self-supervised learning could make for a quite generalizable representation for the re-ID task, better than supervised learning representation trained on a curated and limited dataset.

Besides supervised methods, we also want to compare with unsupervised ones. There are few unsupervised methods that report results in the DG setting. The only one that we find is TrackContrast~\cite{trackcontrast}, which train the representation with a tracklet-wise contrastive learning task on a self-collected video dataset, CampusT.  CampusT contains 480-hour videos, 15.4M person images of around 11K identities. The videos are collected by 24 fixed cameras deployed in a closed campus therefore an identity can be captured by multiple cameras. 
CycAs\_R50 outperforms TrackContrast by an obvious margin, \ie, $+21.3\%$ mAP in Market-1501 and  $+20.6\%$ mAP in DukeMTMC-ReID. 
Moreover, CycAs enjoys advantages in several other aspects: First, TrackContrast relies on pseudo labels annotated by an accurate multi-object tracker,  while CycAs requires no pre-labeling and pre-processing except for cropping person images from the video frames. Second, 
although our LMP-video dataset has similar size to that of CampusT, our collection process is far more easy since most videos are crawled from the Internet. 

Finally, we are curious to know whether advanced neural network architectures with larger capacity could better leverage large-scale training data. To this end, we select a most popular recent Transformer~\cite{vit} based architecture, Swin-Transformer~\cite{swin}, as our backbone model. Results are denoted as CycAs\_Swin in \tabref{tab:exp:dg}. CycAs\_Swin consistently outperforms CycAs\_R50 in all datasets, especially in the most challenging CUHK03 dataset, it improves upon CycAs\_R50 by $+10.6\%$ mAP.  This suggests that advanced architectures like vision transformers could indeed better leverage large-scale training data, and therefore generalizes better on unseen domains. Compared with existing supervised methods considered, the unsupervised CycAs\_Swin shows the best performance in 7 out of the 8 datasets, refreshing state-of-the-art results.

\begin{table*}[tb]

\centering

    \begin{tabular}{l c c c cc cc cc cc  }
    \toprule
           \multirow{2}{*}{Method} & \multirow{2}{*}{Category} & \multirow{2}{*}{Require} & Training &   \multicolumn{2}{c}{Market~\cite{market}}  & \multicolumn{2}{c}{Duke~\cite{duke}} & \multicolumn{2}{c}{CUHK03~\cite{cuhk03}} & \multicolumn{2}{c}{MSMT17~\cite{msmt}} \\
           \cmidrule{5-12}
           &&& Complexity&R1 & mAP & R1 & mAP & R1 & mAP & R1 & mAP \\
           \midrule
         SPGAN~\cite{SPGAN} & UDA & Sup. Pretrain. & $\mathcal{O}(N)$ &   51.5 & 22.8 & 41.1 & 22.3 &- & - & - & - \\
         SPGAN+LMP~\cite{SPGAN} & UDA & Sup. Pretrain. & $\mathcal{O}(N)$&   57.7 & 26.7 &  46.4 & 26.2 &- & - & - & - \\
         TJ-AIDL~\cite{tjaidl} & UDA &  Sup. Pretrain. & $\mathcal{O}(N)$& 58.2 & 26.5 & 44.3 & 23.0  &- & - & - & - \\
         HHL~\cite{HHL}& UDA & Sup. Pretrain. & $\mathcal{O}(N)$& 62.2 & 31.4 & 46.9 & 27.2 & - & - & - & - \\
         ECN~\cite{ecn} & UDA & Sup. Pretrain. & $\mathcal{O}(N)$& 75.1 & 43.0 & 63.3 & 40.4 & - & - &  30.2 & 10.2 \\

         TAUDL~\cite{taudl} & Tracklet & MOT & $\mathcal{O}(N)$&  63.7 & 41.2 & 61.7 & 43.5 & 44.7 & 31.2 & 28.4 & 12.5 \\
         UTAL~\cite{utal} & Tracklet & MOT & $\mathcal{O}(N)$ & 69.2 & 46.2 & 62.3 & 44.6 &  \textbf{56.3} & \textbf{42.3} &  31.4 & 13.1 \\
         UGA~\cite{uga} & Tracklet & MOT & $\mathcal{O}(N)$ & {87.2} & {70.3} &  \emph{75.0} & \emph{53.3} &  - & - &  \emph{49.5} &\emph{21.7} \\
         
         PUL~\cite{pul}& Clustering & Sup. Pretrain.  & $\mathcal{O}(N\log N) \sim \mathcal{O}(N^2)$ & 44.7 & 20.1 &30.4 &16.4  & - & - & - & - \\
         CAMEL~\cite{camel}& Clustering & Sup. Pretrain. & $\mathcal{O}(N\log N)\sim \mathcal{O}(N^2)$ & 54.5 & 26.3& - & - & 39.4 &- &- & -\\
         BUC~\cite{buc}& Clustering & None & $\mathcal{O}(N\log N) \sim \mathcal{O}(N^2)$ & 66.2 & 38.3 & 47.4 & 27.5 & - & - & - & - \\
         CDS~\cite{CDS}& Clustering & Sup. Pretrain.& $\mathcal{O}(N\log N) \sim \mathcal{O}(N^2)$  & 71.6 & 39.9 &67.2 & 42.7 & - & - & - & - \\
         SpCL~\cite{spcl} & Clustering & None & $\mathcal{O}(N\log N) \sim \mathcal{O}(N^2)$& \textbf{88.1} & \textbf{73.1}  & - & - & - & - & 42.3 & 19.1 \\

         \textbf{CycAs$^\texttt{asy}$}& Self-Sup & None & $\mathcal{O}(N)$ & \emph{84.8} & \emph{64.8} & \textbf{77.9} & \textbf{60.1} &  \emph{47.4} & \emph{41.0} & \textbf{50.1} &\textbf{26.7} \\
        \midrule
         \textbf{CycAs$^\texttt{sym}$}& - & None & $\mathcal{O}(N)$ & 88.1 & 71.8 &  79.7 & 62.7&56.4 &  49.6 & 61.8 &  36.2 \\
         IDE & Supervised & Label & $\mathcal{O}(N)$ & 89.2 & 73.9 & 80.0 & 63.1 & 54.2 & 47.2 & 60.2 & 33.4 \\
         \bottomrule

    \end{tabular}
    \caption{Comparison  with state-of-the-art methods on four standard  datasets. Note all the methods starts from a ImageNet pretrained model. The requirement \emph{Pretain} and \emph{None} refers to whether pretraining on labeled re-ID datasets is needed. Training complexity refers to space and time complexity of the training process \wrt the dataset size $N$.
    CycAs$^\texttt{asy}$ is our method, and CycAs$^\texttt{sym}$ refers to an upper bound of our method.}
    \label{tab:exp:unsup}

\end{table*}

\subsection{Canonical Unsupervised re-ID}
\label{sec:exp:unsup}
In this section, we compare with prior arts in the canonical setting of unsupervised re-ID. In this setting, training and test data are sampled from the same domain, so the comparison is mainly on the representation learning capacity without consideration of domain gap.
 We evaluate on four mainstream datasets, \ie, Market-1501~\cite{market}, CUHK03~\cite{cuhk03}, DukeMTMC-ReID~\cite{duke}, and MSMT17~\cite{msmt}. 
 All the datasets do not provide video-level data so we are not able to train CycAs from scratch.
 Instead, since camera indices are annotated in these datasets, we can simulate the required intra- / inter-video sampling for CycAs. Following existing practice \cite{taudl,utal,uga}, we assume all images per ID per camera are drawn from a single tracklet.
Consider a mini-batch with batch size $B$, to mimic the intra- / inter-video sampling, we first randomly sample $B/2$ identities.
For intra-video sampling, one tracklet is sampled for each of these identities; then, two bounding boxes are sampled within each tracklet.
For inter-video sampling, two tracklets from different cameras are sampled for each of the $B/2$ identities; then, one image is sampled from each tracklet. 
Different from previous experiments in the DG re-ID setting, since there is no domain gap between training and test sets, we do not add  IN layers and employ a standard ResNet-50 for fair comparison. Also, parameters of BN layers are not freezed during training. Cross-batch memory is not used because we find it hardly helps on such small-scale training sets.
Results are shown in \tabref{tab:exp:unsup}

\textbf{Performance upper bound analysis.} It is noticeable that the simulated data are absolutely symmetric, \ie, $\tau = 1$, according to the sampling strategy described above.  The performance under this setting can be seen as an upper bound of the proposed method, denoted as CycAs$^\texttt{sym}$. 
For comparison, we implement a supervised baseline (IDE~\cite{market}). We observe that the performance of CycAs$^\texttt{sym}$ is consistently competitive on all the datasets. Compared with IDE, the rank-1 accuracy of CycAs$^\texttt{sym}$ is lower by only $1.1\%$, $0.3\%$ and $1.6\%$ on Market-1501, DukeMTMC-ReID and MSMT17, respectively. 
On CUHK03, CycAs$^\texttt{sym}$ even surpasses IDE by $+2.2\%$. This result partially prove the good alignment between the CycAs task and the objective of re-ID.

\textbf{Comparison with the state of the art.} For fair comparisons, we train CycAs under a practically reasonable asymmetric assumption. 
Specifically, we control intra-sampling symmmetry $\tau_\alpha$ and inter-sampling asymmetry $\tau_\beta$ by replacing a portion of images in $\mathcal{I}_2$ with randomly sampled images from irrelevant identities.
We fix $\tau_\alpha=0.9$ and $\tau_\beta=0.6$, and compare the results (denoted as CycAs$^\texttt{asy}$) with existing unsupervised re-ID approaches. Three categories of existing methods are compared, \ie, unsupervised domain adaptation (UDA) \cite{SPGAN,tjaidl,HHL,ecn}, 
clustering-based methods~\cite{pul,camel,buc,CDS,DGM,RACE}, and tracklet-based methods~\cite{taudl,utal,uga}.
Beside re-ID accuracy, we also compare another dimension, \ie, ease of use, by listing the requirements and training complexities (space and time) of each method in ~\tabref{tab:exp:unsup}. The influence of various  intra-sampling symmmetry $\tau_\alpha$ and inter-sampling asymmetry $\tau_\beta$ will be described in~\secref{sec:exp:ablation} later.

CycAs$^\texttt{asy}$ achieves state-of-the-art accuracy on two larger datasets, \ie, DukeMTMC and MSMT17. The mAP improvement over the second best method~\cite{uga} is $+6.8\%$ and $+5.0\%$ on DukeMTMC and MSMT17, respectively. 
On Market-1501 and CUHK03,  CycAs$^\texttt{asy}$ is very competitive to the best performing methods SpCL~\cite{spcl} and UTAL~\cite{utal}.

Comparing with other unsupervised strategies, CycAs requires less external supervision. For example, UDA methods use a labeled source re-ID dataset, and most clustering-based methods need a pre-trained model for initialization, which also uses external labeled re-ID datasets.
The tracklet-based methods do not require re-ID labels, but require a good tracker to provide good supervision signals.Training such a good tracker also requires external pedestrian labels. 
CycAs learns person representations directly from videos and does not require any external annotation. Its requiring less supervision and competitive accuracy making it potentially a more practical solution for unsupervised re-ID. 
 
 Also note that a recent popular line of works represented by SpCL~\cite{spcl} do not require external annotations, either. They start with an ImageNet pretrained representation, performs clustering to assign pseudo labels to samples and then apply contrastive learning. Due to  accurate pseudo labels, such clustering-based methods perform pretty well and become state-of-the-arts in unsupervised learning. However, we argue that the quadratic training complexity of these methods is an important drawback, which prevent them from being applied to large-scale datsets. 
 By contrast, CycAs does not require clustering and the training complexity is linear to the size of training data. This makes CycAs a more applicable solution to large-scale unsupervised re-ID than clustering methods.

\subsection{Pre-training $\rightarrow$ Fine-tuning}
\label{sec:exp:pretrain}

Serving as pre-train models is a most mainstream usage of current self-supervised representations.  Typically, a model is first trained with self-supervised pretext tasks on a large dataset such as ImageNet~\cite{imagenet} to learn general-purpose representation, and then fine-tuned on the target dataset to learn downstream task related knowledge. Unsurprisingly, such strategy has been found useful in the re-ID task, also. 
In this section, we evaluate how CycAs performs as pre-train models, and compare with three widely-used baselines: (1) supervised representation pre-trained on ImageNet~\cite{imagenet}, (2) self-supervised representation pretrained on ImageNet with MoCo~\cite{moco}, and (3) self-supervised representation pre-trained with MoCo~\cite{moco} on the person-centric LUP~\cite{LUP} dataset. Following the experimental setup in~\cite{LUP}, we employ these pre-trained representations to initialize the backbone network of a state-of-the-art re-ID model, MGN~\cite{mgn}, and then train MGN models on given datasets. We experiment on four datasets: Market-1501, DukeMTMC, CUHK03 and MSMT. Results are shown in~\tabref{tab:exp:pretrain}. Recent state-of-the-art results on the four datsets considered are also listed for comparison.

\begin{table}[]
    \centering
     \caption{Results of fine-tuning a state-of-the-art re-ID model, MGN, with different pre-train models. IN refers to ImageNet supervised pre-trained model. MoCo refers to self-supervised pre-trained model trained on ImageNet with MoCo~\cite{moco}. LUP refers to self-supervised pre-trained model trained on the person-centric LUP~\cite{LUP} dataset with MoCo. }
    \begin{tabular}{l c c c c }
         \toprule 
         Method & Market & Duke & CUHK03 & MSMT17 \\
         \midrule
         PCB~\cite{pcb} & 81.6/93.8 & 69.2/83.3 & 57.5/63.7 & - \\
         BOT~\cite{bagoftricks}& 85.9/94.5 & 76.4/86.4 & - & - \\
         DSA~\cite{dsa}&  87.6/95.7 & 74.3/86.2 & 75.2/78.9 & - \\
         ABDNet~\cite{abdnet} &88.3/95.6 &78.6/89.0 & - & 60.8/82.3 \\
         OSNet~\cite{osnet} & 84.9/94.8 & 73.5/88.6 & 67.8/72.3 & 52.9/78.7 \\
         MHN~\cite{mhn} & 85.0/95.1 &77.2/89.1 & 72.4/77.2 &  - \\
         BDB~\cite{bdb} &  86.7/95.3 & 76.0/89.0 & \underline{76.7}/\underline{79.4} & -\\
         SONA~\cite{sona} & 88.8/95.6 & 78.3/89.4 & \textbf{79.2}/\textbf{81.8} & - \\
         GCP~\cite{gcp} & 88.9/95.2 & 78.6/87.9 & 75.6/77.9 & - \\
         ISP~\cite{isp}  & 88.6/95.3 & 80.0/89.6 & 74.1/76.5 & - \\
         GASM~\cite{gasm} &  84.7/95.3 & 74.4/88.3 & - & 52.5/79.5 \\
         \midrule
         IN+MGN &  87.5/95.1 &79.4/89.0& 70.5/71.2& 63.7/85.1 \\
         MoCo+MGN &  88.2/95.3 & 79.5/89.1 &  67.1/67.0 & 62.7/84.3 \\
         LUP+MGN & \underline{91.0}/\underline{96.4} &\underline{82.1}/\underline{91.0} & 74.7/75.4 & \underline{65.7}/\underline{85.5} \\
         CycAs+MGN & \textbf{91.2}/\textbf{96.5} & \textbf{82.7}/\textbf{91.1} & 76.3/76.9 & \textbf{65.8}/\textbf{86.1} \\
         
         \bottomrule
    \end{tabular}
   
    \label{tab:exp:pretrain}
\end{table}

CycAs consistently outperforms other pre-trained models on all the four datsets. Comparing between IN+MGN and MoCo+MGN, we find the self-supervised method MoCo performs as well as supervised learning. Comparing between MoCo+MGN and LUP+MGN, it is clear that pre-training on the person-centric datasets LUP  significantly boosts re-ID performance, showing the importance of minimizing domain gaps between pre-training and target datsets. The pre-training data of CycAs, LMP-video,  can be seen as drawn from the same data domain as the LUP dataset since we use the same (but a partial) video list. The consistent improvements of CycAs suggest the proposed cycle association pre-text task could serve as a better objective than MoCo, for learning pre-trained re-ID representation.

A convenient benefit of pre-trained models is that it readily improves  existing models.  We show that by using CycAs pre-trained models for initialization, an MGN model could already outperform most previous strong supervised re-ID models, refreshing state-of-the-art on Market-1501, Duke-MTMC, and MSMT17.

\textbf{Remark: } Though CycAs performs well serving as pre-trained models, pre-training is not the purpose that it is originally designed for. It should be noted that the unique advantage of CycAs distinct from other pre-trained models, is that the learned representation could be used directly for re-ID \emph{without fine-tuning}. We also evaluate previous pre-trained models (MoCo trained on LUP) in an experimental setting that fine-tuning is not performed, 
and find they fail in such a setting. Results will be described in~\secref{sec:exp:scalability}.

\subsection{Key Factors to Study}
\label{sec:exp:ablation}

\subsubsection{Scalability to Various Data Size}
\label{sec:exp:scalability}
In this section, we show our method has good scalability to large-scale, noisy training data. We mainly discuss along three dimensions: First, how the performance of the representation, in terms of re-ID accuracy, improves with more data; Second, how the training time grows with more data; And finally, how the maximum memory required in training grows with more data.
If not specified, we use the LMP-youtube subset for training, so only intra-video sampling is applicable for CycAs. We split the training set into 11 subsets with increasing sizes $2^k\times10^4, k=1,2,\cdots, 11$, in terms of image numbers,  and evaluate along the above dimensions on these subsets.
For comparison, we also experiment with two existing, prevalent unsupervised representations trained on the same data:

\begin{enumerate}
\item {SpCL}~\cite{spcl}: A state-of-the-art unsupervised re-ID method. DBSCAN~\cite{dbscan} clustering algorithm is performed before every training epoch, assigning pseudo labels to samples. 
Based on the pseudo labels,  contrastive learning is performed for this epoch. The clustering and training process iterates until convergence. 

\item {MoCo-V3}~\cite{moco}: A classic self-supervised  method for general-purpose representation learning. Currently it is also used in the re-ID task, \eg, in~\cite{LUP}, but mainly serves as pre-training models for supervised fine-tuning. However,  even without fine-tuning, it is natural to expect the learned representation has dicriminative ability to certain extent,  and we are curious to know how good it could be. 
\end{enumerate}

\begin{figure}
    \centering
    \includegraphics[width=\linewidth]{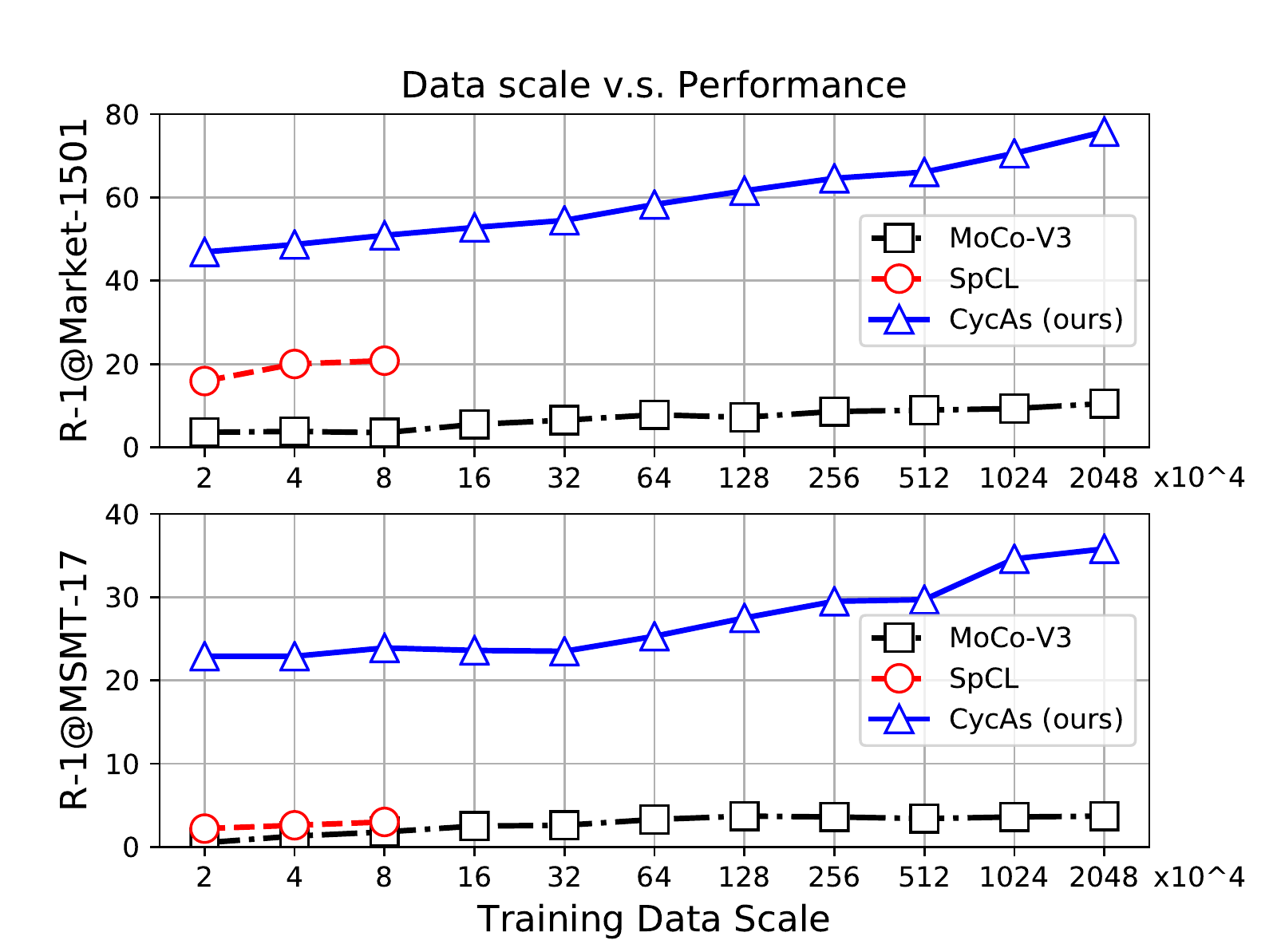}
    \caption{Relationships between training data scale and re-ID accuracy in terms of R-1 on Market-1501~\cite{market} (top) and MSMT-17~\cite{msmt} (bottom). Note that for SpCL, results of more than $8\times10^4$ training images are not applicable, because the  memory consumed exceeds our limit (128G).}
    \label{fig:exp:acc_vs_scale}
\end{figure}

\begin{figure}
    \centering
    \includegraphics[width=\linewidth]{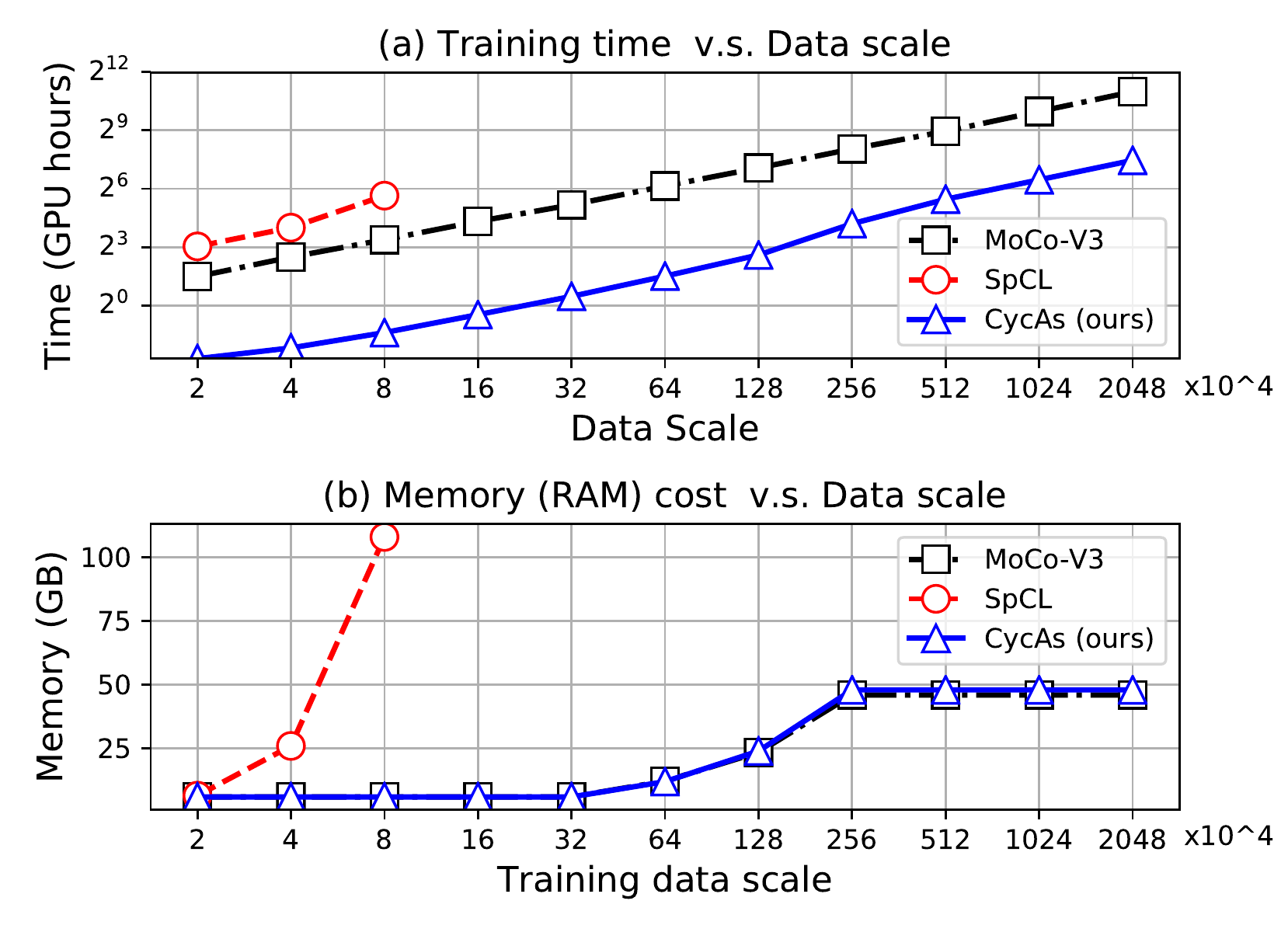}
    \caption{Relationships between training data scale and (a) training time in GPU hours and (b) maximum memory used. Note we only show results of SpCL with training images less than $8\times10^4$, because the  memory consumed exceeds our limit (128G) with more training data.}
    \label{fig:exp:ts_vs_scale}
\end{figure}

Below we detail comparisons along the three  dimensions mentioned. \figref{fig:exp:acc_vs_scale} and \figref{fig:exp:ts_vs_scale} summarize the comparisons.

\textbf{Re-ID accuracy \vs data scale.}
From \figref{fig:exp:acc_vs_scale}, we observe that the re-ID accuracy of the general-purpose SSL representation MoCo-V3~\cite{moco} hardly improves when more data is used for training. The Rank-1 accuracy in Market-1501 improves  merely  by $6.9\%$ when we increase the data scale from $2\times 10^4$ to $2048\times 10^4$, and the performance is rather low even when a large amount of data is used for training ($10.5\%$ Rank-1 when using $2048\times 10^4$ images). This observation is consistent with former attempts in adapting MoCo to the re-ID task~\cite{spcl,LUP}, in which the main reason of the low accuracy is attributed to the instance discrimination pretext task used in MoCo. There is a contradiction between instance discrimination and re-ID: the former aims at assigning a unique representation to each instance, whilst the latter aims at assigning an identical representation to instances belonging to the same identity. Given this contradiction in learning objectives, it is not surprising 
that the MoCo method poorly adapts to the re-ID task. 

The state-of-the-art unsupervised re-ID method, SpCL~\cite{spcl}, shows a moderate scalability to the size of training data. When we increase the data scale from $2\times 10^4$ to $8\times 10^4$, \ie, enlarging the training data by $4$ times, the accuracy improves by $+4.9\%$ (from $15.9\%$ to $20.8\%$). However, we are not able to further investigate the scalability of SpCL when the data scale keeps growing, because the memory consumed by SpCL exceeds our limit when the data scale is larger than $8\times 10^4$.

Finally,  the proposed CycAs method shows stable scalability to the size of training data. Along with the growth of the  training data scale, the re-ID accuracy in both the considered datasets consistently keeps improving, and we find the Rank-1 accuracy roughly grows linear to the logarithm of data scale. 
Two points are worth noting from \figref{fig:exp:acc_vs_scale}: First, in the cases of fewer training samples, CycAs already outperforms MoCo-V3 and SpCL by a large margin; and second, even trained on as many as $2048\times10^4$ samples, we still do not observe a trend in performance saturation from the curve of CycAs. This suggests that the CycAs representation fits well to the re-ID task, and hopefully has great potential if the training set grows even larger.

\textbf{Training time cost \vs data scale.} In~\figref{fig:exp:ts_vs_scale}a we compare how the training time increases with growing data scales. In general, the training time of all the three methods considered grows linearly to the data scale.  Among them, the time cost of the proposed CycAs is the lowest, typically  greater than 10 times faster than  MoCo-V3, and 100 times faster than SpCL. With the largest data scale $2\times 10^7$, the training of CycAs takes  176 GPU hours, which means the training finishes in only 22 hours on $8\times$ Nvidia A100 GPU cards. Note that, theoretically, the training time of SpCL is supposed to increase 
faster than linearly to the data scale, since when the data scale $N$ grows sufficiently large, the clustering algorithm would become the  bottleneck of computation, which has an $O(N\log N)$ to $O(N^2)$ complexity.

\textbf{Training memory cost \vs data scale.} 
In~\figref{fig:exp:ts_vs_scale}b we compare how the memory consumed in training increases with growing data scales. It can be clearly observed that SpCL has a quadratic curve. When the data scale is $8\times 10^4$, the memory used is 108 GB. In our experiments,  we use a machine with 128 GB RAM, therefore when the data scale is greater than $8\times 10^4$, we are not able to run SpCL. We examine the SpCL algorithm and find that the main memory cost comes from several $N^2$ size matrices used to compute ``independence'' and ``compactness'' of clusters. There is no doubt that it is possible to optimize the algorithm so that the memory cost could be hopefully reduced to $O(N\log N)$, but such complexity is still prohibitive when the data scale becomes quite large. By contrast, we show that both MoCo-V3 and the proposed CycAs consumes memory roughly linearly to the data scale.  In fact, the linear growth of memory cost is mainly attributed to the increased number of GPU cards used, since in each GPU we start an independent data loader that consumes RAM. Accordingly, the memory cost could even be constant if we use a fixed number of GPUs. For instance, we use 8 GPUs in all experiments when the data scale varies from $256$ to $2048\times 10^4$, so the curve of memory cost plateaus in this range.

\subsubsection{Data sources \vs Scalability}
\label{sec:exp:dataset_scalability}
\begin{figure}
    \centering
    \includegraphics[width=\linewidth]{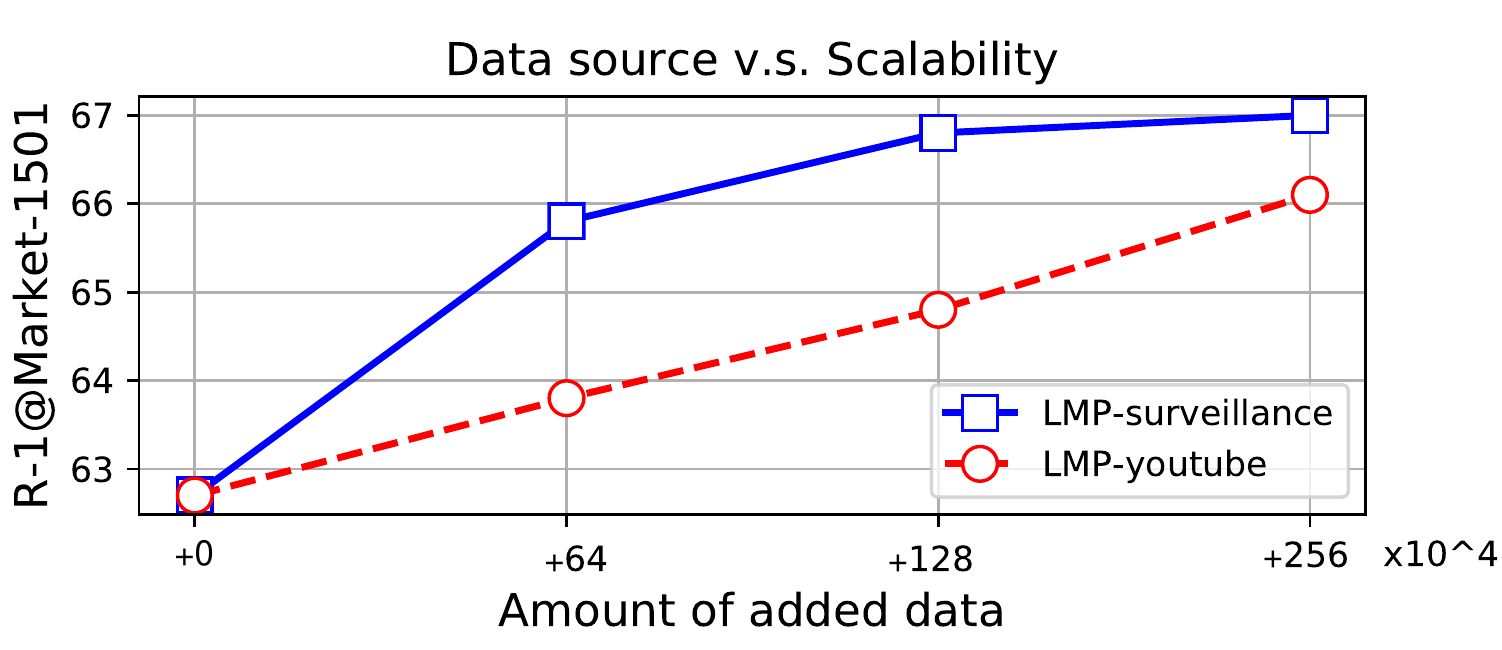}
    \caption{CycAs has different scalability to different data sources. For Youtube videos, performance grows slowly but steadily with more training data; In contrast, for surveillance videos, performance improve faster but soon saturates.}
    \label{fig:scouce_scalability}
\end{figure}

In the above section we discussed scalability of CycAs on the LMP-youtube subset, and below we discuss the characteristic of scalability on the LMP-surveillance subset, which is quite different as pointed in \secref{sec:dataset:summary}. Since training solely on LMP-surveillance does not converge, we start from using $256\times 10^4$ samples from LMP-youtube, and gradually add $64, 128, 256\times 10^4$ samples from LMP-surveillance. For comparison, we also perform another group of experiments by gradually adding $64, 128, 256\times 10^4$ samples from LMP-youtube. Results are shown in \figref{fig:scouce_scalability}. 

Several observations can be made from the figure. First, both surveillance videos and Youtube videos help. 
Second, adding a small amount of surveillance videos is more beneficial than adding an equivalent amount of Youtube videos. This is mainly because LMP-surveillance provides paired videos, so that inter-video sampling is applicable. Due to larger appearance changes, inter-video sampled data help learn a more discriminative representation.
Finally, we do not observe performance saturation when adding more Youtube data, but there is an obvious saturation trend when adding more surveillance data. We speculate the reason is that our surveillance videos are collected from a pair of \emph{fixed} cameras, thus significantly lacks domain diversity.

 \subsubsection{Robustness against different levels of symmetry $\tau$.} 
\begin{figure}[t]
     \centering
     \includegraphics[width=\linewidth]{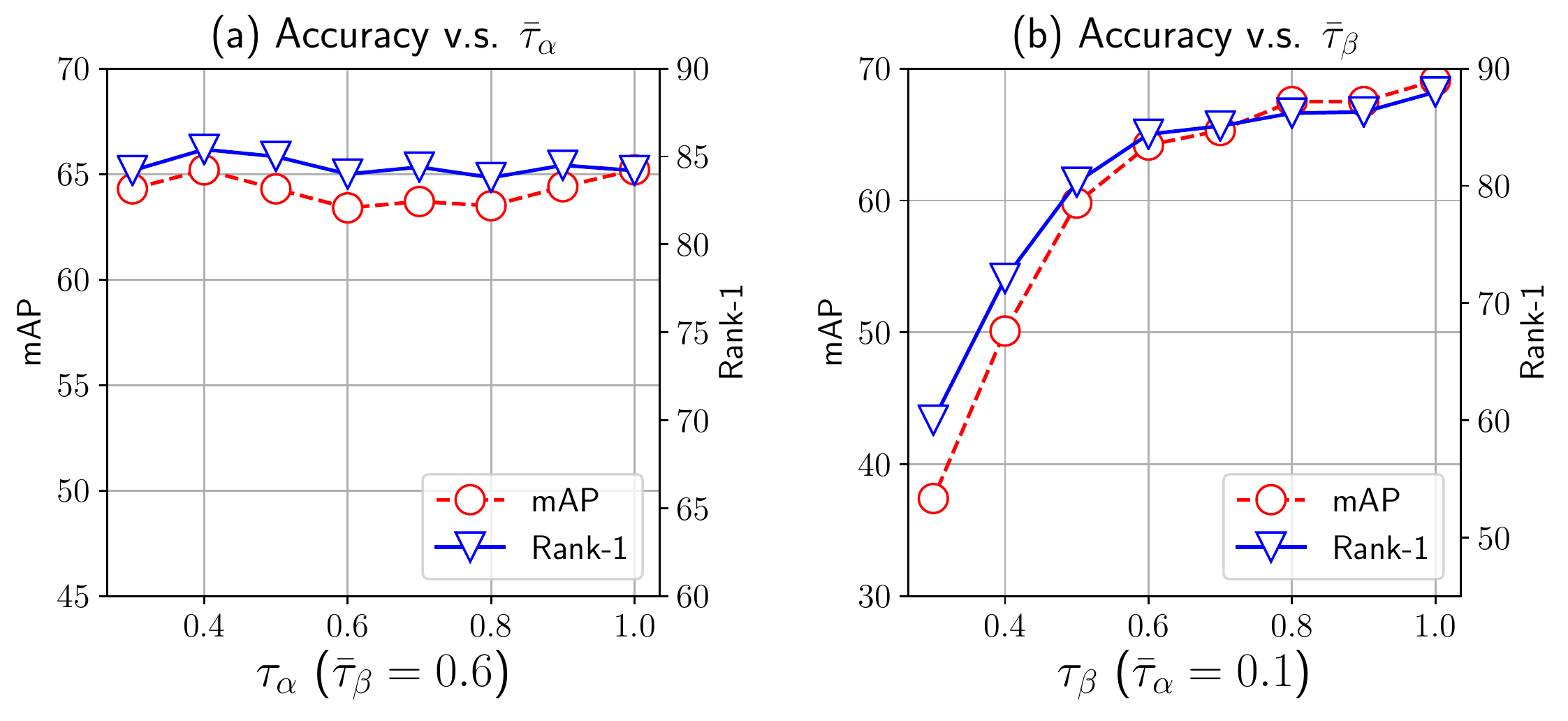}
     \caption{Robustness against different levels of (a) intra-sampled data symmetry $\tau_\alpha$ and (b) inter-sampled data symmetry $\tau_\beta$. For evaluating $\tau_\alpha$, in each mini-batch, we fix $\tau_\beta$ and draw a random $\tau_\alpha$ from $\mathcal{N}(\bar{\tau}_\alpha, 0.01)$, and plot the performance curve \wrt different mean $\bar{\tau}_\alpha$. The impact of $\tau_\beta$ is evaluated in a similar way. }
   \label{fig:exp:symmetry}
 \end{figure}

 To investigate the robustness of CycAs against different levels of symmetry, we experiment with various data symmetry and observe how the ReID accuracy changes. 
 In order to manually control the data symmetry, we employ the labeled Market-1501 dataset for training, and follow the setting in unsupervised re-ID (\secref{sec:exp:unsup}) to simulate asymmetry.
 Specifically, we control intra-sampling symmmetry $\tau_\alpha$ and inter-sampling asymmetry $\tau_\beta$ by replacing a portion of images in $\bm{I}_2$ with randomly sampled images from irrelevant identities.
 To evaluate the impact of $\tau_\alpha$, in each mini-batch, we fix $\tau_\beta$ and draw $\tau_\alpha$ from a gaussian distribution $\mathcal{N}(\bar{\tau}_\alpha, 0.01)$, truncated in range $(0,1)$, and plot the model performance against the mean $\bar{\tau}_\alpha$. The impact of $\tau_\beta$ is evaluated in a similar way. We report results  in Fig. \ref{fig:exp:symmetry}.

Two observations can be made. First, with a moderate fixed value of $\tau_\beta$, \ie, $0.6$ in our case, the model accuracy is robust to a wide range of $\bar{\tau}_\alpha$. For example, in  Fig.~\ref{fig:exp:symmetry}a, the rank-1 accuracy is the same $84.2\%$ when $\bar{\tau}_\alpha$ is set to $1$ and $0.3$, respectively.
Second, the accuracy improves when $\bar{\tau}_\beta$ becomes larger. 
The reason is that re-ID requires a discriminative representation that can well-describe the appearance  of persons even with a large variation in poses.
The knowledge learned from intra-sampled data contributes less to the overall performance. Therefore the final accuracy is less sensitive to $\bar{\tau}_\alpha$. In contrast, learning from inter-sampled data aligns with the objective of re-ID task, therefore the final accuracy is more sensitive to $\bar{\tau}_\beta$.

\textbf{Remarks.} Note that in Fig.~\ref{fig:exp:symmetry}b, the curves drop very slowly when $\tau_\beta =1$ decreases from $1$ to $0.6$. 
This suggests that CycAs has a good ability in handling moderate data asymmetry. 
Such a property is valuable, because in practice we can control $\bar{\tau}_\beta$ in a reasonable range (say from $0.6$ to $0.9$), by deploying the cameras in desired positions. Compared  with manual data annotation, careful deployment of cameras  clearly requires much less effort.

\subsubsection{Hyper-parameters in XBM}
We investigate how  hyper-parameters in XBM affect the final performance. Two hyper-parameters are considered, \ie, the mermory size $M$, and $k$ for the top-$k$ hard negative mining.  \tabref{tab:xbm}  shows results on Market-1501 test set.

\begin{table}[]
    \centering
    \begin{tabular}{l|c c c c c}
    \toprule
         Memory size $M$ & 0  & 16K & 32K & 64K & 128K\\
         \midrule
         
         Rank-1 & 75.5 &78.0 &78.9 & \textbf{80.3} & 80.2 \\
        \bottomrule
    \end{tabular}

    \vspace{0.3cm}
    
        \begin{tabular}{l|c c c c c}
    \toprule
         \# hard negative $k$ & 1 & 5 & 10 & 50 & 100 \\
         \midrule
         
         Rank-1 & 78.8 & 79.6 & \textbf{80.3} & 79.5 & 80.0\\
        \bottomrule
    \end{tabular}
    \caption{Ablation on XBM hyper-parameters.}
    
    \label{tab:xbm}
\end{table}

For the memory size, We find the best selection is $M=64K$. Compared with $M=0$, which means XBM is not applied, $M=64K$ brings $+4.8\%$ improvement in Rank-1. Further expanding the memory size does not bring improvements. Therefore, we select the best combination of $M=64K$ in most experiments. Because XBM stores compact features, the memory cost is rather small: it only costs 128 MB extra GPU memory. 

Regarding the number of sampled hard negative $k$, when we increase $k$ from $1$ to $10$, performance improves. However, if keep increasing $k$ when $k$ is larger than 10, no further improvements are observed. Therefore, we set $k=10$ by default.

\section{Conclusion}
In this work, we show that with the help of an adequate self-supervised learning algorithm,  simply scaling up training data significantly improves the generalization capability of re-ID representations. Our contributions are two fold: First, we construct a multi-person video dataset, LMP-video, that is large in scale and diverse in data distribution. Second, we propose CycAs, a novel self-supervised learning pre-text task that employs the  cycle consistency after  forward and backward association as free supervision signals. The CycAs method shows good scalability to different training data sizes. Trained on the large-scale LMP-video dataset, our best results refresh state-of-the-art in Domain Generalization re-ID, and also perform favorably against best existing results in unsupervised re-ID and re-ID pre-training settings.
The fact that CycAs surpasses supervised methods trained with limited data  shows scalable self-supervised learning is a promising path towards  generalizable person re-ID.


%


\ifCLASSOPTIONcaptionsoff
  \newpage
\fi




\clearpage
\bibliographystyle{bib/IEEEtran}
\bibliography{bib/egbib}
\clearpage
\appendices
\section{Adaptive Softmax temporature}
Consider two vectors with different sizes, $\bm{v}=(1,0.5)^\top$ and $\bm{u}=(1,0.5, 0.5)^\top$. Let $\bm{\sigma}$ be the softmax operation, then $\bm{\sigma(v}) = (0.62,0.38)^\top$ and $\bm{\sigma(u}) = (0.45, 0.27, 0.27)^\top$. 
The \emph{Soft-Max} operation, as we observe, has different levels of softening ability on inputs with different sizes.
The Max value in a longer vector is less highlighted, or maxed, and vice versa. To alleviate this problem and stabelize the training, we let the softmax temperature be adaptive to the varying input size, so that for input vectors of different sizes, the max values in them are equally highlighted.

To fulfill this goal, here we give a simple method as an effective estimation of $T$, which is adaptive to the size $K$. Assuming for a specific instance, there is one positive match and $(K-1)$ negative matches. The cosine similarity between the positive pair is $p$ while those between the negative pairs are $n$. For a specific difference between positive / negative scores $\epsilon = p - n$, We expect the difference after the softmax operation is still a constant $\delta$, regardless of size $K$
\begin{equation}
    \frac{e^{Tp} - e^{Tn}}{e^{Tp} + (K-1)e^{Tn}} = \frac{e^{T\epsilon} - 1}{e^{T\epsilon} + (K-1)} = \delta
\end{equation}
which yields
\begin{equation}
T = \frac{1}{\epsilon} \log \left[ \frac{\delta (K-1) + 1}{1-\delta} \right]
\end{equation}
The value of $\delta$ can be simply fixed. We use $\delta=0.5$, so that $T = \frac{1}{\epsilon} \log (K+1)$.
There is still one hyperparameter $\epsilon$ to be tuned, so the number of hyperparameters remains the same. However, tuning $\epsilon$ is much easier than directly tuning $T$, since $\epsilon$ is bounded in $[-1,1]$ while $T$ is unbounded.  Also, we can use a fixed $\epsilon$, as the final  temperature is adaptive to size $K$.

\ifCLASSOPTIONcompsoc
  \section*{Acknowledgments}
\else
  \section*{Acknowledgment}
\fi

This work is supported by the state key development program in 14th Five-Year
under Grant No. 2021YFF0602103, 2021YFF0602102, 2021QY1702. We also thank for the research
fund under Grant No. 2019GQG0001 from the Institute for Guo Qiang, Tsinghua University.

\end{document}